\documentclass[10pt,twocolumn,letterpaper]{article}

\usepackage{iccv}
\usepackage{times}
\usepackage{epsfig}
\usepackage{graphicx}
\usepackage{amsmath}
\usepackage{amssymb}

\usepackage{booktabs}
\usepackage{xcolor}
\usepackage{colortbl}
\usepackage{multirow}
\usepackage[ruled,vlined,noend]{algorithm2e}

\usepackage{comment}
\usepackage{caption}
\usepackage{algorithmic}
\newcommand{\taun}{\tau_\mathsfit{S}}
\usepackage{float}
\usepackage{enumitem}
\usepackage{microtype}

\usepackage{amsmath,amsfonts,bm}

\def\eqref#1{equation~\ref{#1}}

\def\1{\bm{1}}

\def\rvp{{\mathbf{p}}}

\def\va{{\bm{a}}}
\def\vb{{\bm{b}}}
\def\vc{{\bm{c}}}

\def\vg{{\bm{g}}}

\def\vo{{\bm{o}}}
\def\vp{{\bm{p}}}
\def\vq{{\bm{q}}}
\def\vr{{\bm{r}}}

\def\vu{{\bm{u}}}
\def\vv{{\bm{v}}}

\def\vx{{\bm{x}}}
\def\vy{{\bm{y}}}
\def\vz{{\bm{z}}}

\def\mS{{\bm{S}}}

\DeclareMathAlphabet{\mathsfit}{\encodingdefault}{\sfdefault}{m}{sl}
\SetMathAlphabet{\mathsfit}{bold}{\encodingdefault}{\sfdefault}{bx}{n}

\def\sC{{\mathbb{C}}}
\def\sD{{\mathbb{D}}}

\def\sR{{\mathbb{R}}}

\def\sX{{\mathbb{X}}}
\def\sY{{\mathbb{Y}}}

\newcommand{\E}{\mathbb{E}}

\DeclareMathOperator*{\argmax}{arg\,max}
\DeclareMathOperator*{\argmin}{arg\,min}

\newcommand\citep[1]{\cite{#1}}
\newcommand\citet[1]{\cite{#1}}
\usepackage{multicol}

\usepackage[accsupp]{axessibility}

\usepackage[pagebackref=true,breaklinks=true,letterpaper=true,colorlinks,bookmarks=false]{hyperref}

 \iccvfinalcopy 

\def\iccvPaperID{2210} 

\ificcvfinal\fi

\begin{document}

\title{Practical Relative Order Attack in Deep Ranking}

\author{
Mo Zhou$^1$~~
Le Wang$^{1\ast}$~~
Zhenxing Niu$^2$~~
Qilin Zhang$^3$~~
Yinghui Xu$^2$~~
Nanning Zheng$^1$~~
Gang Hua$^4$\\
$^1$Xi'an Jiaotong University ~~
$^2$Alibaba Group ~~
$^3$HERE Technologies ~~
$^4$Wormpex AI Research\vspace{-0.3em}\\
{\tt\scriptsize \{cdluminate,zhenxingniu,samqzhang,ganghua\}@gmail.com}
{\tt\scriptsize lewang@xjtu.edu.cn}
{\tt\scriptsize renji.xyh@taobao.com}
{\tt\scriptsize nnzheng@mail.xjtu.edu.cn}\vspace{-0.8em}
}

\maketitle
\ificcvfinal\thispagestyle{empty}\fi

{\renewcommand{\thefootnote}{*}
\footnotetext{Corresponding author.}}

\newcommand{\blue}{black}
\newif\ifcvpr
\cvprtrue
\newif\ifsupp
\suppfalse

\begin{abstract}

	Recent studies unveil the vulnerabilities of deep ranking models,
	where an imperceptible perturbation can trigger dramatic changes in the
	ranking result.
	While previous attempts focus on manipulating \emph{absolute ranks} of
	certain candidates, the possibility of adjusting their \emph{relative
	order} remains under-explored.
	\textcolor{\blue}{
	In this paper, we formulate a new adversarial attack against deep ranking
	systems, \ie, the \emph{Order Attack}, which covertly alters the
	\emph{relative order} among a selected
	set of candidates according to an attacker-specified permutation,
	with limited interference to other unrelated candidates.}
	\textcolor{\blue}{
	Specifically, it is formulated as a triplet-style loss imposing an inequality
	chain reflecting the specified permutation. However, direct optimization
	of such white-box objective is infeasible in a real-world
	attack scenario due to various black-box limitations.}
	\textcolor{\blue}{
	To cope with them, we propose a
	\emph{Short-range Ranking Correlation} metric as a surrogate objective for
	black-box \emph{Order Attack} to approximate the white-box method.}
	\textcolor{\blue}{
	The \emph{Order Attack} is evaluated on the
	Fashion-MNIST and Stanford-Online-Products datasets under both white-box and
	black-box threat models.
	The black-box attack is also successfully implemented on a major e-commerce platform.}
	\textcolor{\blue}{
	Comprehensive experimental evaluations demonstrate
	the effectiveness of the proposed methods, revealing a new type of ranking model 
	vulnerability.}

\end{abstract}

\section{Introduction}
\label{sec:intro}

Thanks to the widespread applications of deep neural networks~\citep{alexnet,resnet} in
 the learning-to-rank tasks~\citep{imagesim2,facenet}, deep ranking algorithms have
 witnessed significant progress, but unfortunately they have also inherited the
 long-standing adversarial vulnerabilities~\cite{l-bfgs} of neural networks. 
Considering the ``search by image'' application for example, an imperceptible adversarial
perturbation to the query image is sufficient to intentionally alter the ranking results
of candidate images. 
Typically, such adversarial examples can be designed to cause the ranking
model to ``misrank''~\citep{advrank-ut2,universalret} (\ie, rank items
incorrectly),
or purposefully raise or lower the ranks of selected candidates
~\citep{advrank}.

\ifcvpr
\begin{figure*}[t]
\centering
	\includegraphics[width=1.0\linewidth]{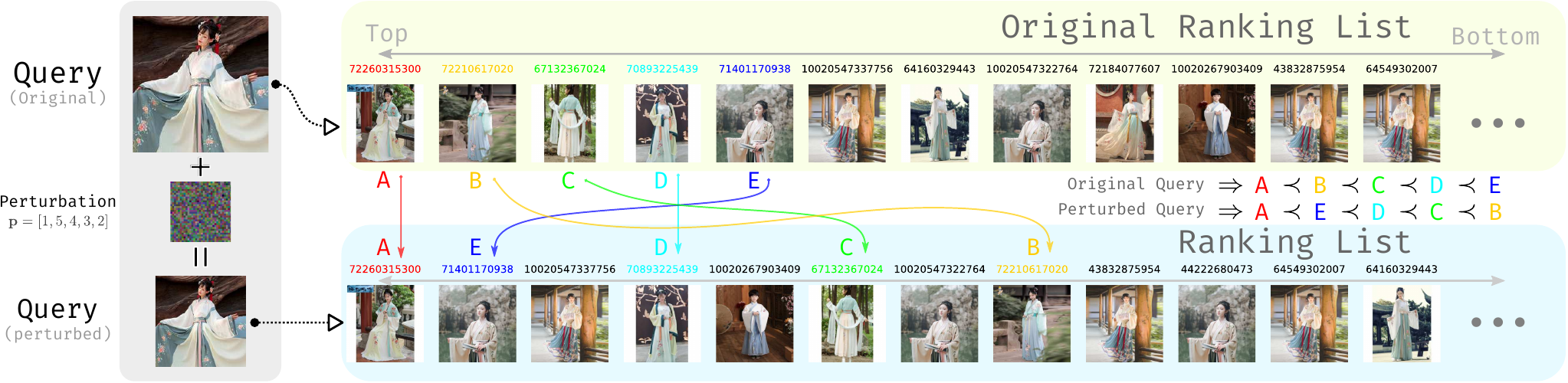}
	\caption{Showcase of a practical \emph{Order Attack} (OA) against ``JD SnapShop'',
	a major online retailing e-commerce platform.
	The query image is ``Han Chinese clothing''.
	Numbers atop candidate images are Stock Keep Unit (SKU) IDs.
	}
\label{fig:advorder}
\end{figure*}
\fi

Since ``misranking'' can be interpreted as deliberately lowering the ranks of well-matching
candidates, previous attacks on ranking models unanimously focus on
changing the \emph{absolute ranks} of a set of candidates, while neglecting the manipulation of \emph{relative order} among them.
\textcolor{\blue}{
However, an altered \emph{relative order} can be disruptive in some applications,
such as impacting sales on e-commerce platforms powered by content-based
image retrieval~\citep{cbir}, where
potential customers attempt to find merchandise via image search.}

\textcolor{\blue}{
As shown in Fig.~\ref{fig:advorder}, an attacker may want to adversarially perturb
the query image and thus change the \emph{relative order}
among products A, B, C, D, and E into $\text{A}\prec
\text{E}\prec\text{D}\prec\text{C}\prec\text{B}$ in the search-by-image result.
The sales of a product closely correlates to its Click-Through Rate (CTR),
while the CTR can be significantly influenced by its ranking position~\cite{chen2012position,pin2011stochastic}
(which also decides the product pagination on the client side).
Hence, subtle changes in the \emph{relative order} of searches can be sufficient
to alter CTR and impact the actual and relative sales among A to E.}

\textcolor{\blue}{
Such vulnerability in a commercial platform may be exploited in a
malfeasant business competition among the top-ranked products,
\eg, via promoting fraudulent web pages containing adversarial example product
images generated in advance.
Specifically, the attack of changing \emph{relative order} does not aim to incur
significant changes in the
\emph{absolute ranks} of the selected candidates (\eg, moving from list bottom to top),
but it intentionally changes the \emph{relative order} of them subtly without
introducing conspicuous abnormality.
Whereas such goal cannot be implemented by \emph{absolute rank} attacks such as
\cite{advrank}, the \emph{relative order} vulnerability may justify and motivate
a more robust and fair ranking model.}


Specifically, we propose the \emph{Order Attack} (OA), a new adversarial
attack problem in deep ranking. Given a query image $\vq\in[0,1]^D$, a set of
selected candidates $\sC=\{\vc_1,\vc_2,\ldots,\vc_k\}$, and a predefined
permutation vector $\rvp=[p_1,p_2,\ldots,p_k]$, \emph{Order Attack} aims to
find an imperceptible perturbation $\vr$ ($\|\vr\|_\infty \leqslant
\varepsilon$ and $\tilde{\vq}=\vq+\vr\in [0,1]^D$), so that $\tilde{\vq}$ as
the adversarial query can convert the \emph{relative order} of the selected candidates
into $\vc_{p_1} \prec \vc_{p_2} \prec \cdots \prec \vc_{p_k}$.  For example, a
successful OA with $\rvp=[1,5,4,3,2]$ will result in $\vc_1 \prec \vc_5 \prec
\vc_4\prec \vc_3\prec \vc_2$, as shown in Fig.~\ref{fig:advorder}.


~

To implement OA, we first assume the
\emph{white-box} threat model (\ie, the ranking model details, \emph{incl.} the
gradient are accessible to the attacker). 
Recall that a conventional deep ranking
model~\citep{imagesim2,ladder,facenet,nonbindml} maps the query and candidates
onto a common embedding space, and determines the ranking list according to the
pairwise similarity between the query and these candidates.
Thus, OA can be formulated as the optimization of a triplet-style loss function
based on the inequality chain representing the desired \emph{relative order},
which simultaneously adjusts the similarity scores between the query and the
selected candidates. Additionally, a semantics-preserving penalty term
\citep{advrank} is also included to limit conspicuous changes in ranking positions.
Finally, the overall loss function can be optimized with gradient methods
such as PGD~\citep{madry} to find the adversarial example.

\textcolor{\blue}{
However, in a real-world \emph{black-box} attack scenario, practical limitations
(\eg, gradient inaccessibility) invalidate the proposed method.
To accommodate them and make OA practical,
we propose a ``Short-range Ranking Correlation'' (SRC) metric to measure the
alignment between a desired permutation and the actual ranking result returned
to clients by counting concordant and discordant pairs,
as a practical approximation of the proposed triplet-style
white-box loss.
Though non-differentiable, SRC can be used as a surrogate objective for black-box
OA and optimized by an appropriate black-box optimizer, to achieve
similar effect as the white-box OA.
SRC can also be used as a performance metric for the white-box method,
as it gracefully degenerates into Kendall's ranking correlation~\citep{kendall} in 
white-box scenario.}


To validate the white-box and black-box OA, we conduct
comprehensive experiments on Fashion-MNIST and Stanford-Online-Product
datasets.
To illustrate the viability of the black-box OA in practice, we also
showcase successful attacks against the ``JD SnapShop''~\citep{jd}, a
major retailing e-commerce platform based on content-based image retrieval.
\textcolor{\blue}{
Extensive quantitative and qualitative evaluations illustrate the effectiveness
of the proposed OA, and reveals a new type of ranking model vulnerability.
}


To the best of our knowledge, this is the first work that tampers the
\emph{relative order} in deep ranking.  We believe our contributions include, 
%
%
\textbf{(1)} the formulation of \emph{Order Attack} (OA), a new adversarial
attack that covertly alters the \emph{relative order} among selected candidates;
\textbf{(2)} a triplet-style loss for ideal-case white-box OA;
\textbf{(3)} a \emph{Short-range Ranking Correlation} (SRC) metric as a surrogate
objective approximating the triplet-style loss for practical black-box OA;
\textbf{(4)} extensive evaluations of OA including a successful demonstration on a
major online retailing e-commerce platform.

\ifcvpr
\else
\begin{figure}[t]
\centering
	\includegraphics[width=1.0\linewidth]{advorder-crop.pdf}
	\caption{Showcase of a practical \emph{Order Attack} (OA) against ``JD SnapShop'',
	a major online retailing e-commerce platform.
	Numbers atop candidate images are 
	Stock Keep Unit (SKU) IDs.
	}
\label{fig:advorder}
\fi

\section{Related Works}

\textbf{Adversarial Attack}. Szegedy \etal~\citet{l-bfgs} find the DNN
classifiers susceptible to imperceptible adversarial perturbations, which
leads to misclassification.
This attracted research interest among the community, as shown
by subsequent works on adversarial attacks and defenses~\cite{benchmarking,autoattack,adaptive}.
In particular, the attacks can be categorized into several groups:
{(1)} White-box attack, which assumes the model details including the gradient
are fully accessible~\citep{fgsm,i-fgsm,madry,deepfool,cw,obfuscated,phy-synth,apgd}.
Of these methods, PGD~\cite{madry} is the most popular one; 
{(2)} Transfer-based attack,
which is based on the transferability of adversarial examples~\citep{mi-fgsm,di-fgsm,ti-fgsm}.
Such attack typically transfers adversarial examples found from a locally trained
substitute model onto another model.
%
{(3)} Score-based attack, which only depends on the soft classification labels,
\ie, the logit values~\citep{nes-atk,spsa-atk,nattack,zoo,square}.
Notably, \citet{nes-atk} proposes a black-box threat model for classification
that is similar to our black-box ranking threat model;
{(4)} Decision-based attack, a type of attack
that requires the least amount of information from the model,
\ie, the hard label (one-hot vector)~\citep{boundaryattack,hsja,opt,evolutionary,bayesian,qeba}.
All these extensive adversarial attacks unanimously focus on classification,
which means they are not directly suitable for ranking scenarios.

\begin{figure}[t]
	\includegraphics[width=1.0\linewidth]{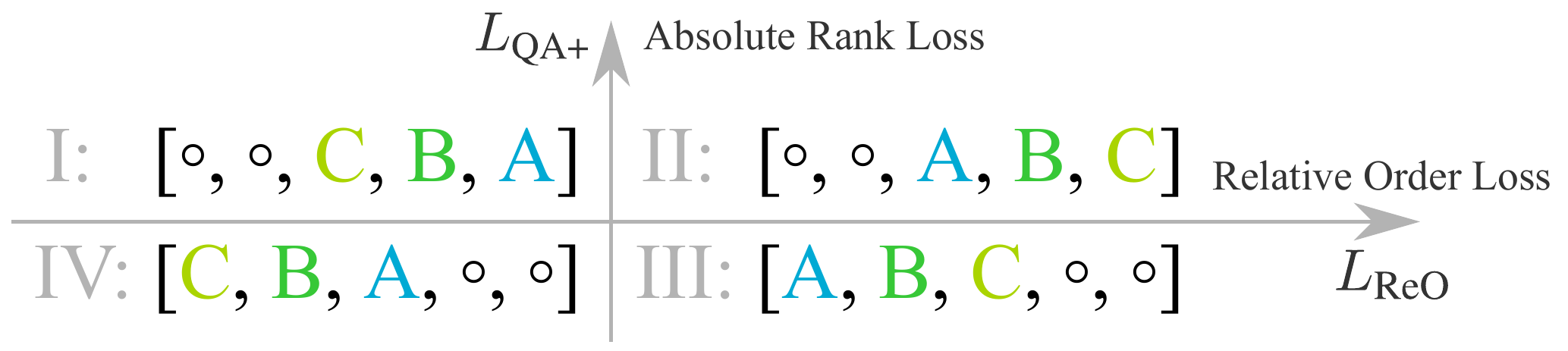}
	\caption{Relative order attack \emph{v.s.} absolute rank attack.}
	\label{fig:reo-qa-orth}
\end{figure}

\textbf{Adversarial Ranking.}
In applications such as web retrieval, documents may be promoted in rankings by intentional manipulation~\cite{rankrobust}.
Likewise, the existence of aforementioned works inspired attacks against deep
ranking, 
but it is still insufficiently explored.
In light of distinct purposes, ranking attacks can be divided into
\emph{absolute rank} attacks and \emph{relative order} attacks.
Most \emph{absolute rank} attacks attempt to induce random
``misranking''~\citep{flowertower,universalret,advrank-ut2,hamming,
zhao2019unsupervised,learn-to-misrank,ofda,metric2,advdpqn,qair}.
Some other attacks against ranking model aim to incur purposeful changes in \emph{absolute rank},
\ie, to raise or lower the ranks of specific candidates~\citep{metric1,advrank}.
%
On the contrary, the \emph{relative order} attacks remains under-explored.
And this is the first work that tampers the \emph{relative order} in deep ranking.

%
As shown in Fig.~\ref{fig:reo-qa-orth}, \emph{relative order} is ``orthogonal''
to \emph{absolute rank}.
Let ``$\circ$'' denote any uninterested candidate. 
Suppose the selected candidates $\sC$ and permutation $\rvp$ are [A, B, C] and $[3,2,1]$,
respectively.
The exemplar \emph{absolute rank} loss $L_\text{QA+}$~\cite{advrank} is ignorant to the difference
in \emph{relative order}
comparing I and II, or IV and III.
The \emph{relative order} loss $L_\text{ReO}$ proposed in this paper is ignorant
to the difference in \emph{absolute rank} comparing I and IV, or II and III.
\textcolor{\blue}{Although focusing on different aspects of ranking,}
the two types of loss functions can be combined and jointly optimized.


\section{Adversarial Order Attack}
\label{sec:3}

Typically, a deep ranking model is built upon deep metric
learning~\citep{imagesim2,facenet,ladder,nonbindml,revisitingdml}.
Given a query $\vq$ and a set of candidates $\sC=\{\vc_1,\vc_2,\ldots,\vc_k\}$
selected from database $\sD$ ($\sC\subset \sD$), a deep ranking model $f$
evaluates the distance between every pair of query and candidate, \ie,
$f:\mathcal{I}\times\mathcal{I}\mapsto \sR$ where $\mathcal{I}=[0,1]^D$.
Thus, by comparing all the pairwise distances
$\{f(\vq,\vc_i)|i=1,2,\ldots,k\}$, the whole candidate set can be ranked with
respect to the given query.
For instance, the model outputs the ranking list $\vc_1\prec \vc_2\prec \cdots
\prec \vc_k$ if it determines $f(\vq,\vc_1)<f(\vq,\vc_2)<\cdots<f(\vq,\vc_k)$.

Based on these, \emph{Order Attack} (OA) aims to find an imperceptible
perturbation $\vr$ ($\|\vr\|_\infty \leqslant \varepsilon$ and
$\tilde{\vq}=\vq+\vr\in \mathcal{I}$), so that $\tilde{\vq}$ as the adversarial
query can convert the \emph{relative order} of the selected candidates into
$\vc_{p_1} \prec \vc_{p_2} \prec \cdots \prec \vc_{p_k}$, where
$\rvp=[p_1,p_2,\ldots,p_k]$ is a permutation vector predefined by the attacker.
In particular, we assume that the attacker is inclined to select the candidate
set $\sC$ from the top-$N$ ranked candidates $\sX$,
as the ranking lists returned to the clients are usually ``truncated''
(\ie, only the top-ranked candidates will be shown).
We call the length $N$ ($N\geqslant k$) of the ``truncated'' list
as a ``visible range''.
The white-box and black-box OA will be discussed in Sec.\ref{sec:white} and
Sec.\ref{sec:black} respectively.  For sake of brevity, we let
$\Omega_\vq=\{\vr|~\vq+\vr\in \mathcal{I},~\|\vr\|_\infty \leqslant \varepsilon
\}$.

\subsection{Triplet-style Loss Function for White-Box OA}
\label{sec:white}

During the training process, a typical deep ranking
model $f$ involves a triplet (anchor $\vq$, positive $\vc_\text{p}$,
negative $\vc_\text{n}$) in each iteration. In order to rank $\vc_\text{p}$ ahead of
$\vc_\text{n}$, the model is penalized when
$f(\vq,\vc_\text{p})+\gamma<f(\vq,\vc_\text{n})$ does not hold.
This inequality can be reformulated exploiting the form of a hinge loss~\cite{hinge}, resulting in the
triplet ranking loss function~\cite{facenet} $L_\text{triplet}(\vq,\vc_\text{p},\vc_\text{n})=
[\gamma + f(\vq,\vc_\text{p}) - f(\vq,\vc_\text{n})]^+$, where
$[\cdot]^+=\max(0,\cdot)$, and $\gamma$ denotes the margin hyper-parameter.

Inspired by this, to implement the OA, we decompose the inequality chain prescribed by
the predefined permutation vector $\rvp$, namely
$f(\tilde{\vq},\vc_{p_1}) {<} f(\tilde{\vq},\vc_{p_2}) {<} \cdots{<}f(\tilde{\vq},\vc_{p_k})$
into $(^k_2) {=} k(k{-}1)/2$ inequalities, \ie, 
$f(\tilde{\vq},\vc_{p_i}) < f(\tilde{\vq},\vc_{p_j})$, $i,j{=}1,2,\ldots,k,~i{<}j$.
Reformulation of these inequalities into the hinge loss form leads to
the \emph{relative order} loss,
\begin{equation} L_{\text{ReO}} (\tilde{\vq};\sC,\rvp) =\sum_{i=1}^k
\sum_{j=i}^k \big[f(\tilde{\vq},\vc_{p_i}) - f(\tilde{\vq},\vc_{p_j}) \big]^+.
\label{eq:reo} \end{equation}
Subsequently, given this loss function, the OA can be cast as a constrained optimization
problem,
\begin{equation}
\vr^\ast=\argmin_{\vr\in\Omega_\vq}L_{\text{ReO}}(\vq+\vr;\sC,\rvp),
\label{eq:oaopt} \end{equation}
which can be solved with first-order-gradient-based methods such as
Projected Gradient Descent (PGD)~\citep{madry}, \ie,
%
\begin{equation} \vr_{t+1}=\text{Clip}_{\Omega_\vq}\big\{ \vr_t - \eta
\text{sign}\big[\nabla_\vr L_\text{ReO}(\tilde{\vq};\sC,\rvp)\big] \big\},
\label{eq:pgd} \end{equation}
where $\eta$ is the PGD step size, and $\vr_0$ is initialized as a zero vector.
PGD stops at a predefined maximum iteration $T$, and the final $\vr_T$
is the desired adversarial perturbation.

It is worth noting that query image semantics
can be drastically changed even with a very slight perturbation~\citep{advrank}.
\textcolor{\blue}{
As a result, candidates $\sC$ are prone to be excluded from the topmost part of ranking,
and become invisible when the ranking result is ``truncated''.}
To mitigate such side effect, we follow \citet{advrank} and introduce a
semantics-preserving term $L_\text{QA+}(\tilde{\vq},\sC)$ to keep $\sC$ within
the topmost part of the ranking by raising their \emph{absolute ranks},
\ie, to keep $\vc\in \sC$ ranked ahead of other candidates.
Finally, the \emph{relative order} loss term $L_\text{ReO}(\cdot)$ and the \emph{absolute
rank} loss term $L_\text{QA+}(\cdot)$ are combined to form the complete 
white-box OA loss $L_\text{OA}$, 
\begin{equation} L_{\text{OA}}(\tilde{\vq};\sC,\rvp) =
L_{\text{ReO}}(\tilde{\vq};\sC,\rvp) + \xi L_{\text{QA+}}(\tilde{\vq},\sC),
\label{eq:ra} \end{equation}
where $\xi$ is a positive constant balancing factor between the  \emph{relative
order} and \emph{absolute rank} goals.


Despite the formulation of Eq.~(\ref{eq:ra}), an ideal $\tilde{\vq}$ that
fully satisfy the desired relative order of $\sC$ does not necessarily
exist.  Consider a Euclidean embedding space, where candidates
$\vc_1,\vc_2,\vc_3$ lie consecutively on a straight line. It is impossible to find a query
embedding that leads to $\vc_1\prec \vc_3 \prec \vc_2$.
That indicates the compatibility between the specified relative order and the
factual geometric relations of the candidate embeddings
affects the performance upper-bound of OA.
In cases like this,
our algorithm can still find an inexact solution that satisfies as many inequalities as
possible to approximate the specified \emph{relative order}.
In light of this, Kendall's ranking correlation $\tau$~\cite{kendall} between the
specified relative order and the real ranking order appears to be a more reasonable
performance metric than the success rate for OA.


\subsection{Short-range Ranking Correlation}
\label{sec:black}

\textcolor{\blue}{
A concrete triplet-style implementation of OA is present in Sec.~\ref{sec:white},
but it is infeasible in a real-world attack scenario.
In particular, multiple challenges are present for black-box OA, including
(1) \emph{Gradient inaccessibility}. The gradient of the loss \emph{w.r.t} the
input is inaccessible, as the network architecture and parameters are unknown;
(2) \emph{Lack of similarity (or distance) scores}. Exact similarity
scores rarely appear in the truncated ranking results;
(3) \emph{Truncated ranking results}. In practice, a ranking system only
presents the top-$N$ ranking results to the clients;
(4) \emph{Limited query budget}. Repeated, intensive queries within a
short time frame may be identified as threats, \eg, Denial of Service (DoS)
attack.  Therefore, it is preferable to construct 
adversarial examples within a reasonable amount of queries.
These restrictions collectively invalidate the triplet-style method.
}

\textcolor{\blue}{
To address these challenges, we present the ``Short-range Ranking Correlation'' (SRC; denoted as
$\taun$) metric, a practical approximation of the $L_\text{OA}$ (Eq.~\ref{eq:ra})
as a surrogate loss function for black-box OA.
%
}

Specifically, to calculate $\taun$ given $\sC$, $\rvp$ and the top-$N$ retrieved
candidates $\sX$ \emph{w.r.t} query $\tilde{\vq}$, we first initialize a
$(k\times k)$-shaped zero matrix $\mS$, and permute $\sC$ into
$\sC_\rvp=\{\vc_{p_1},\vc_{p_2},\ldots,\vc_{p_k}\}$.
Assuming $\forall \vc_i, \vc_j \in \sC_\rvp$ ($i>j, i\neq j$)
exist in $\sX$, we define $(\vc_i,\vc_j)$ as a \emph{concordant} pair
as long as $\text{R}_{\sC_\rvp}(\vc_i)$ and $\text{R}_{\sX}(\vc_i)$
are simultaneously greater or smaller than
$\text{R}_{\sC_\rvp}(\vc_j)$ and $\text{R}_{\sX}(\vc_j)$, respectively,
where $\text{R}_{\sX}(\vc_i)$ denotes the integer rank value of $\vc_i$ in $\sX$,
\ie, $\text{R}_{\sX}(\vc_i) := \arg_m \{\vc_i = \vx_m\}$.
Otherwise, $(\vc_i,\vc_j)$ is defined as a \emph{discordant} pair.
\textcolor{\blue}{
Namely a \emph{concordant} matches a part of the specified permutation,
and could result in a zero loss term in Eq.~\ref{eq:reo},
while a \emph{discordant} pair does not match, and could result in a positive
loss term in Eq.~\ref{eq:reo}.
Thus, in order to approximate the \emph{relative order} loss $L_\text{ReO}$
(Eq.~\ref{eq:reo}),
a \emph{concordant} pair and a \emph{discordant} pair will be assigned
a score of $S_{i,j}=+1$ (as reward) and $S_{i,j}=-1$ (as penalty), respectively.
}
Apart from that, when $\vc_i$ or $\vc_j$ does not exist in $\sX$,
$S_{i,j}$ will be directly assigned with an ``out-of-range'' penalty $-1$,
\textcolor{\blue}{
which approximates the semantics-preserving term in Eq.~\ref{eq:ra}.
Finally, after comparing the ordinal relationships of every pair of candidates
and assigning penalty values in $\mS$,
the average score of the lower triangular of $\mS$
excluding the diagonal is the $\taun$, as summarized in Algo.~\ref{algo:taun}.}

\begin{algorithm}[t]
	\SetAlgoLined
	\KwIn{Selected candidates $\sC=\{\vc_1,\vc_2,\ldots,\vc_k\}$,
	permutation vector $\rvp=[p_1,p_2,\ldots,p_k]$,
	top-$N$ retrieval $\sX=\{\vx_1,\vx_2,\ldots,\vx_N\}$ for $\tilde{\vq}$.
	Note that $\sC\subset\sD$, $\sX\subset\sD$, and $N\geqslant k$.
	}
	\KwOut{SRC coefficient $\taun$.}
	Permute candidates as $\sC_\rvp=\{\vc_{p_1},\vc_{p_2},\ldots,\vc_{p_k}\}$\;
	Initialize score matrix $\mS=\mathbf{0}$ of size ${k\times k}$\;
	\SetInd{0.67em}{0.67em}
	\For{$i \gets 1,2,\ldots,k $}{
		\For{$j \gets 1,2,\ldots,i-1$}{
			\uIf{\upshape $\vc_i \notin \sX$ \footnotemark~ or $\vc_j \notin \sX$
			}{
				$S_{i,j}=-1$ \hfill\tcp{out-of-range}
			}\ElseIf{\upshape
				{\small \big[$\text{R}_{\sC_\rvp}(\vc_i) {>} \text{R}_{\sC_\rvp}(\vc_j)$}
				and {\small $\text{R}_{\sX}(\vc_i) {>} \text{R}_{\sX}(\vc_j)$\big]}
				or
				{\small \big[$\text{R}_{\sC_\rvp}(\vc_i) {<} \text{R}_{\sC_\rvp}(\vc_j)$}
				and {\small $\text{R}_{\sX}(\vc_i) < \text{R}_{\sX}(\vc_j)$\big]}
			}{
				$S_{i,j}=+1$ \hfill\tcp{~ concordant}
			}\ElseIf{\upshape
				{\small \big[$\text{R}_{\sC_\rvp}(\vc_i) {>} \text{R}_{\sC_\rvp}(\vc_j)$}
				and {\small $\text{R}_{\sX}(\vc_i) {<} \text{R}_{\sX}(\vc_j)$\big]}
				or
				{\small \big[$\text{R}_{\sC_\rvp}(\vc_i) {<} \text{R}_{\sC_\rvp}(\vc_j)$}
				and {\small $\text{R}_{\sX}(\vc_i) {>} \text{R}_{\sX}(\vc_j)$\big]}
			}{
				$S_{i,j}=-1$ \hfill\tcp{~ discordant}
			}
		}
	}
	\Return{$\taun = \sum_{i,j}S_{i,j} / (^k_2)$}
	\caption{Short-range Ranking Correlation $\taun$.}
	\label{algo:taun}
\end{algorithm}
\footnotetext{$\nexists m\in\{1,2,\ldots,N\}$ so that $\vc_i = \vx_m$.}

\textcolor{\blue}{
The value of $\taun$ $\in [-1, 1]$ reflects the real ranking order's
alignment to the order specified by $\rvp$,
where a semantics-preserving penalty is spontaneously incorporated.
%
When the specified order is fully satisfied, \ie, any pair of $\vc_i$
and $\vc_j$ is \emph{concordant}, and none of the elements
in $\sC$ disappear from $\sX$, $\taun$ will be $1$.
In contrast, when every candidate
pair is \emph{discordant} or absent from the top-$N$ result $\sX$, $\taun$
will be $-1$.
Overall, $(\taun + 1)/2$ percent of the candidate pairs are \emph{concordant},
and the rest are \emph{discordant} or ``out-of-range''.%
}

\textcolor{\blue}{
Maximization of $\taun$ leads to the best alignment to the specified permutation,
as discordant pairs will be turned into concordant pairs, while
maintaining the presence of $\sC$ within the top-$N$ visible range.
Thus, although non-differentiable, 
the $\taun$ metric can be used as a practical surrogate objective for black-box OA,
\ie,  $\vr^\ast = \argmax_{\vr\in \Omega_\vq} \taun(\tilde{\vq};\sC,\rvp)$,
which achieves a very similar effect to the white-box OA.}

\textcolor{\blue}{
Particularly, when $\forall \vc\in\sC$ exists in $\sX$, $\taun$ degenerates
into Kendall's $\tau$~\cite{kendall} between $\rvp$ and the permutation of $\sC$ in $\sX$.
Namely, $\taun$ also degenerates gracefully to $\tau$ in the white-box scenario
because the whole ranking is visible.
However, $\tau$ is inapplicable on truncated ranking results.}

\textcolor{\blue}{
$\taun$ does not rely on any gradient or any similarity score, and
can adapt to truncated ranking results.
When optimized with an efficient black-box optimizer (\eg, NES~\cite{nes-atk}),
the limited query budget can also be efficiently leveraged.
%
%
Since all the black-box challenges listed at the beginning of this section
are handled, it is practical to perform
black-box OA by optimizing $\taun$ in real-world applications.
}

%

\begin{table*}[t]
\centering
\resizebox{1.0\linewidth}{!}{
\setlength{\tabcolsep}{0.8em}%
\begin{tabular}{c|ccccc|ccccc|ccccc}
	\toprule
 & \multicolumn{5}{c|}{$k=5$} & \multicolumn{5}{c|}{$k=10$} & \multicolumn{5}{c}{$k=25$}\tabularnewline
\hline 
$\varepsilon$ & 0 & $\frac{2}{255}$ & $\frac{4}{255}$ & $\frac{8}{255}$ & $\frac{16}{255}$ & 0 & $\frac{2}{255}$ & $\frac{4}{255}$ & $\frac{8}{255}$ & $\frac{16}{255}$ & 0 & $\frac{2}{255}$ & $\frac{4}{255}$ & $\frac{8}{255}$ & $\frac{16}{255}$\tabularnewline
\midrule
\rowcolor{lime!10}\multicolumn{16}{c}{Fashion-MNIST \qquad $N=\infty$}\tabularnewline
$\taun$ & 0.000 & 0.286 & 0.412 & 0.548 & 0.599 & 0.000 & 0.184 & 0.282 & 0.362 & 0.399 & 0.000 & 0.063 & 0.108 & 0.136 & 0.149\tabularnewline
mR & 2.0 & 4.5 & 9.1 & 12.7 & 13.4 & 4.5 & 7.4 & 10.9 & 15.2 & 17.4 & 12.0 & 16.1 & 17.6 & 18.9 & 19.4\tabularnewline
	\midrule
\rowcolor{cyan!10}\multicolumn{16}{c}{Stanford Online Products \qquad $N=\infty$}\tabularnewline
$\taun$ & 0.000 & 0.396 & 0.448 & 0.476 & 0.481 & 0.000 & 0.263 & 0.348 & 0.387 & 0.398 & 0.000 & 0.125 & 0.169 & 0.193 & 0.200\tabularnewline
mR & 2.0 & 5.6 & 4.9 & 4.2 & 4.1 & 4.5 & 12.4 & 11.2 & 9.9 & 9.6 & 12.0 & 31.2 & 28.2 & 25.5 & 25.4\tabularnewline
\bottomrule
\end{tabular}
}
\vspace{-0.7em}
\caption{White-box order attack on Fashion-MNIST and SOP datasets with various settings.}
\label{tab:white}
\end{table*}

\begin{table}[t]
\centering%
\resizebox{1.0\linewidth}{!}{
\setlength{\tabcolsep}{0.3em}%
\begin{tabular}{c|cccccccccc}
	\toprule
$\xi$ & $0$ & $10^{-1}$ & $10^{0}$ & $10^{1}$ & $10^{2}$ & $10^{3}$ & $10^{4}$ & $10^{5}$ & $10^{6}$ & $10^{7}$\tabularnewline
\midrule
	\rowcolor{lime!10}\multicolumn{11}{c}{Fashion-MNIST \qquad $k=5$, $N=\infty$, $\varepsilon=4/255$}\tabularnewline
$\taun$ & \underline{0.561} & 0.467 & \multicolumn{1}{c|}{0.451} & \multicolumn{1}{c|}{\textit{0.412}} & 0.274 & 0.052 & 0.043 & 0.012 & 0.007 & 0.002\tabularnewline
mR & 27.2 & 22.7 & \multicolumn{1}{c|}{18.3} & \multicolumn{1}{c|}{\textit{9.1}} & 4.9 & 3.2 & 2.8 & 2.7 & 2.7 & 2.7\tabularnewline
\midrule
	\rowcolor{cyan!10}\multicolumn{11}{c}{Stanford Online Products \qquad $k=5$, $N=\infty$, $\varepsilon=4/255$}\tabularnewline
$\taun$ & \underline{0.932} & 0.658 & 0.640 & 0.634 & \multicolumn{1}{c|}{0.596} & \multicolumn{1}{c|}{\textit{0.448}} & 0.165 & 0.092 & 0.013 & 0.001\tabularnewline
mR & 973.9 & 89.8 & 48.1 & 22.4 & \multicolumn{1}{c|}{7.5} & \multicolumn{1}{c|}{\textit{4.9}} & 2.9 & 2.8 & 2.8 & 2.7\tabularnewline
\bottomrule
\end{tabular}
}
\vspace{-0.7em}
\caption{Searching for balancing factor $\xi$ on both datasets.}
\label{tab:xi}
\end{table}


\section{Experiments}
\label{sec:4}


To evaluate the white-box and black-box OA,
we conduct experiments on the
Fashion-MNIST~\citep{fashion} and the Stanford-Online-Products (SOP)~\citep{sop}
datasets which comprise images of retail commodity.
Firstly, we train a CNN with $2$-convolution-$1$-fully-connected network
on Fashion-MNIST, and a ResNet-18~\citep{resnet} without the last fully-connected layer on SOP
following \citet{advrank} that focuses on the \emph{absolute rank} attack.
Then we perform OA with the corresponding test set as the candidate database $\sD$.
Additionally, we also qualitatively evaluate black-box OA on ``JD SnapShop''~\citep{jd} to further illustrate its efficacy. 
In our experiments, the value of rank function $\text{R}_\sX(\cdot)$ starts from $0$, 
\ie, the $k$-th ranked candidate has the rank value of $k-1$.

\textbf{Selection of $\sC$ and $\rvp$.}
As discussed, we assume that the attacker is inclined to select
the candidate set $\sC$ from the top-$N$ ranked candidates given the visible range limit. 
For simplicity, we only investigate the $(k,N)$-OA, \ie, OA with the
top-$k$-within-top-$N$ ($k\leq N$) candidates selected as as $\sC$.
It is representative because an OA problem with some candidates randomly selected
from the top-$k$ results as $\sC$ is a sub-problem of $(k,N)$-OA.
Namely, our attack will be effective for any selection of $\sC$ as long as
the $(k,N)$-OA is effective.
For white-box OA, we conduct experiments with $N{=}\infty$,
and $k{\in}\{5,10,25\}$.
For black-box attack, we conduct experiments with $N{=}\{\infty,50,k\}$, and $k{=}\{5,10,25\}$.
A random permutation vector $\rvp$ is specified for each query.

\textbf{Evaluation Metric.}
Since $\taun$ is equivalent to $\tau$ when $N=\infty$,
we use $\taun$ as the performance metric for both white-box and black-box OA.
Specifically, in each experiment, we conduct $\mathsfit{T}=10^4$ times of OA attack.
In each attack, we randomly draw a sample from $\sD$ as the query $\vq$.
In the end, we report the average $\taun$ over the $\mathsfit{T}$ trials.
Also, when $N=\infty$, we additionally calculate the mean rank of the candidate set $\sC$
(demoted as ``mR'', which equals $[\sum^k_i \text{R}_{\sX}(\vc_i)]/k$),
and report the average mean rank over the $\mathsfit{T}$ attacks.
Larger $\taun$ value and smaller mR value are preferable.

\textbf{Parameter Settings.}
We set the perturbation budget as $\varepsilon\in \{\frac{2}{255}, \frac{4}{255},
\frac{8}{255}, \frac{16}{255} \}$ following \cite{i-fgsm} for both white-box and black-box attacks.
The query budget $Q$ is set to $1.0\times 10^3$.
For white-box OA, the PGD step size $\eta$ is set to $\frac{1}{255}$,
the PGD step number to $24$. The balancing parameter $\xi$ is set as $10^1$
and $10^3$ for Fashion-MNIST and SOP respectively.
The learning rates of black-box optimizer NES~\cite{nes-atk} and SPSA~\cite{spsa-atk}
are both set to $2/255$.
See supplementary for more details of the black-box optimizers.

\textbf{Search Space Dimension Reduction.}
As a widely adopted trick, dimension reduction of the adversarial perturbation
search space has been reported effective in~\citep{benchmarking,zoo,bayesian,qeba}.
Likewise, we empirically reduce the space to $(3\times 32\times 32)$
for black-box OA on Stanford-Online-Products dataset and ``JD SnapShop''.
In fact, a significant performance drop in $\taun$ can be observed
without this trick.

\subsection{White-Box Order Attack Experiments}
\label{sec:expwhite}

\begin{figure*}[t]
	\centering
	\includegraphics[width=1.0\linewidth]{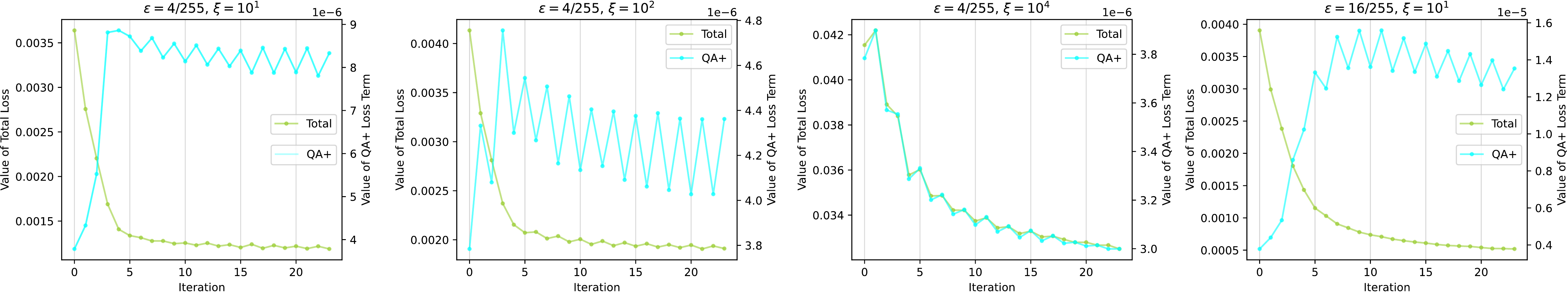}

	\includegraphics[width=1.0\linewidth]{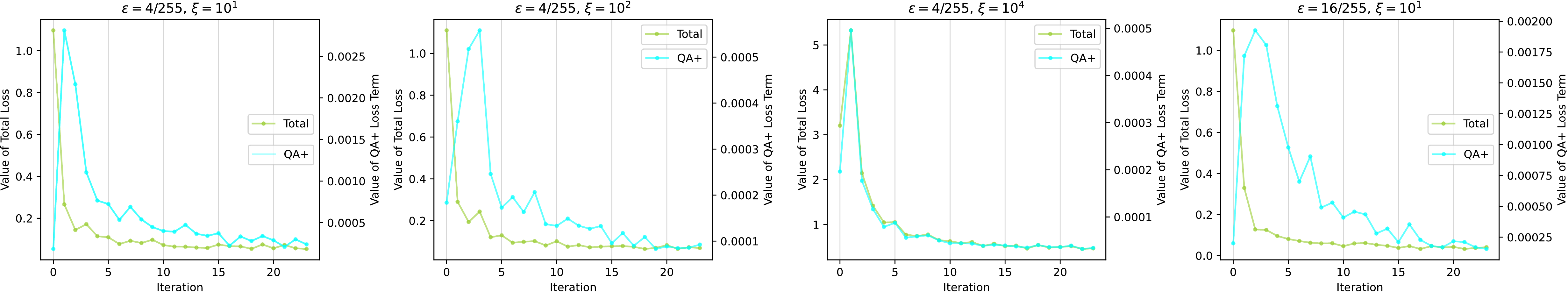}\vspace{-3pt}
	\caption{Curves of total loss $L_\text{OA}$ (left y-axis) and the $L_\text{QA+}$ term (right y-axis) during the
	optimization procedures under different $\varepsilon$ and $\xi$ settings.
	The first row is for Fashion-MNIST, while the second row is for SOP dataset. }
	\label{fig:wloss}
\end{figure*}

The first batch of the experiments is carried out on the Fashion-MNIST dataset, 
as shown in the upper part of Tab.~\ref{tab:white}.
With the original query image ($\varepsilon=0$), the expected
$\taun$ performance of $(5,\infty)$-OA is $0.000$,
and the $\sC$ retains their original ranks as the mR equals $2.0$.
With a $\varepsilon=2/255$ adversarial perturbation budget, our OA
achieves $\taun=0.286$, which means on average $64.3\%$
\footnote{
Solution of $\frac{(n_\text{concordant}{-}n_\text{discordant})}{(^k_2)}{=}0.286$; $\frac{(n_\text{concordant}{+}n_\text{discordant})}{(^k_2)}{=}1.0$.
}
of the inequalities reflecting the specified permutations are satisfied
by the adversarial examples.
Meanwhile, the mR changes from $2.0$ to $4.5$,
due to adversarial perturbation can move the query embedding off its
original position~\cite{advrank} while seeking for a higher $\taun$.
Nevertheless, the mR value of $4.5$ indicates that the $\sC$ are still
kept visible in the topmost part of the ranking result by the loss term $L_\text{QA+}(\cdot)$.
With larger perturbation budget $\varepsilon$, 
the $\taun$ metric increases accordingly, \eg, $\taun$ reaches
$0.599$ when $\varepsilon=16/255$, which means nearly $80\%$ of the inequalities
are satisfied.
Likewise, the experimental results on SOP are available
in the lower part of Tab.~\ref{tab:white}, which also demonstrate the effectiveness
of our method under different settings. 

Besides, we note that different balancing parameter $\xi$ for $L_\text{QA+}(\cdot)$ leads to
distinct results, as shown in Tab.~\ref{tab:xi}.
We conduct $(5,\infty)$-OA
with $\varepsilon=4/255$ with different $\xi$ values ranging from $0$ to $10^7$ on
both datasets. Evidently, a larger $\xi$ leads to
a better (smaller) mR value, but meanwhile a worse $\taun$ as the
weighted $L_\text{QA+}(\cdot)$ term dominates the total loss.
There is a trade-off between the $\taun$ and mR, which is
effectively controlled by the tunable constant parameter $\xi$.
Hence, we empirically set $\xi$ as $10^1$ and $10^3$ for Fashion-MNIST and SOP
respectively, in order to keep the mR of most experiments in Tab.~\ref{tab:white}
below a sensible value, \ie, $50/2$.
\parskip 1pt


Additionally, Tab.~\ref{tab:white} reveals that the mR trends w.r.t. 
$\varepsilon$ on the two datasets differ.
%
To investigate this counter-intuitive phenomenon, we plot loss curves in Fig.~\ref{fig:wloss}.
In the $\varepsilon=4/255$, $\xi=10$ case for Fashion-MNIST, the total loss decreases
but the $L_\text{QA+}$ surges at the beginning and then plateaus.
After increasing $\xi$ to $10^2$, the $L_\text{QA+}$ rises more smoothly.
The curve eventually decreases at $\xi=10^4$, along with a small mR and
a notable penalty on $\taun$ as a result.
Besides, the ``sawtooth-shaped'' $L_\text{QA+}$ curves also indicate that
the $L_\text{ReO}$ term is optimized while sacrificing the mR as a side-effect
at the even steps, while the optimizer turns to optimize $L_\text{QA+}$ at
the odd steps due to the semantics-preserving penalty, causing a slight increase
in $L_\text{ReO}$.
These figures indicate that optimizing $L_\text{ReO}$ without sacrificing
$L_\text{QA+}$ is difficult.
%
Moreover, perturbation budget is irrelevant.
Comparing the first and the fourth sub-figures, we find
a larger budget ($\varepsilon{=}\frac{16}{255}$) unhelpful in reducing optimization
difficulty as the $L_\text{QA+}$ curve still soars and plateaus.
Based on these cues, we speculate that the different curve patterns
of mR stem from the optimization difficulty due to a \emph{fixed
PGD step size} that cannot be smaller\footnote{Every element of perturbation should be an integral multiple of $1/255$.},
and \emph{different dataset properties}.

The intra-class variance of the simple Fashion-MNIST dataset is smaller
than that of SOP, which means
sample embeddings of the same class are densely clustered.
As each update can change the $\sX$ drastically,
it is difficult to adjust the query embedding position in a dense area
with a fixed PGD step for a higher $\taun$ without significantly disorganizing
the ranking list (hence a lower mR).
%
In contrast, a larger intra-class variance of the SOP dataset makes
$L_\text{QA+}$ easier to be maintained, as shown in the 2nd row of Fig.~\ref{fig:wloss}.
%


\ifcvpr
\begin{table*}[t]
\else
\begin{table}[t]
\fi
\centering
\resizebox{1.0\linewidth}{!}{
\setlength{\tabcolsep}{0.3em}%
\begin{tabular}{c|cccc|cccc|cccc}
\toprule
	\multirow{2}{*}{\textbf{Algorithm}} & \multicolumn{4}{c|}{$k=5$} & \multicolumn{4}{c|}{$k=10$} & \multicolumn{4}{c}{$k=25$}\tabularnewline
\cline{2-13} \cline{3-13} \cline{4-13} \cline{5-13} \cline{6-13} \cline{7-13} \cline{8-13} \cline{9-13} \cline{10-13} \cline{11-13} \cline{12-13} \cline{13-13}
 & \multicolumn{1}{c|}{$\varepsilon=\frac{2}{255}$} & \multicolumn{1}{c|}{$\varepsilon=\frac{4}{255}$} & \multicolumn{1}{c|}{$\varepsilon=\frac{8}{255}$} & $\varepsilon=\frac{16}{255}$ & \multicolumn{1}{c|}{$\varepsilon=\frac{2}{255}$} & \multicolumn{1}{c|}{$\varepsilon=\frac{4}{255}$} & \multicolumn{1}{c|}{$\varepsilon=\frac{8}{255}$} & $\varepsilon=\frac{16}{255}$ & \multicolumn{1}{c|}{$\varepsilon=\frac{2}{255}$} & \multicolumn{1}{c|}{$\varepsilon=\frac{4}{255}$} & \multicolumn{1}{c|}{$\varepsilon=\frac{8}{255}$} & $\varepsilon=\frac{16}{255}$\tabularnewline
\hline
None & 0.0, 2.0 & 0.0, 2.0 & 0.0, 2.0 & 0.0, 2.0 & 0.0, 4.5 & 0.0, 4.5 & 0.0, 4.5 & 0.0, 4.5 & 0.0, 12.0 & 0.0, 12.0 & 0.0, 12.0 & 0.0, 12.0\tabularnewline
\midrule
	\rowcolor{lime!10}\multicolumn{13}{c}{Fasion-MNIST \qquad $N=\infty$}\tabularnewline
Rand & 0.211, 2.1 & 0.309, 2.3 & 0.425, 3.0 & 0.508, 7.7 & 0.172, 4.6 & 0.242, 5.0 & 0.322, 6.4 & 0.392, 12.7 & 0.084, 12.3 & 0.123, 13.1 & 0.173, 15.8 & 0.218, 25.8\tabularnewline
Beta & 0.241, 2.1 & 0.360, 2.6 & 0.478, 4.6 & 0.580, 19.3 & 0.210, 4.8 & 0.323, 5.7 & 0.430, 9.6 & 0.510, 30.3 & 0.102, 12.4 & 0.163, 13.8 & 0.237, 19.7 & 0.291, 42.7\tabularnewline
PSO & 0.265, 2.1 & 0.381, 2.3 & 0.477, 4.4 & 0.580, 21.1 & 0.239, 4.8 & 0.337, 5.7 & 0.424, 9.7 & 0.484, 34.0 & 0.131, 12.7 & 0.190, 14.6 & 0.248, 21.7 & 0.286, 54.2\tabularnewline
NES & 0.297, 2.3 & \textbf{0.416, 3.1} & \textbf{0.520, 8.7} & \textbf{0.630, 46.3} & \textbf{0.261, 5.0} & 0.377, 6.6 & 0.473, 14.3 & 0.518, 55.6 & \textbf{0.142, 13.0} & 0.217, 15.9 & 0.286, 28.3 & 0.312, 74.3\tabularnewline
SPSA & \textbf{0.300, 2.3} & 0.407, 3.2 & 0.465, 7.1 & 0.492, 16.3 & 0.249, 5.0 & \textbf{0.400, 6.6} & \textbf{0.507, 12.8} & \textbf{0.558, 27.5} & 0.135, 12.9 & \textbf{0.236, 16.3} & \textbf{0.319, 27.1} & \textbf{0.363, 46.4}\tabularnewline
\midrule
	\rowcolor{lime!20}\multicolumn{13}{c}{Fashion-MNIST \qquad $N=50$}\tabularnewline
Rand & 0.207 & 0.316 & 0.424 & 0.501 & 0.167 & 0.242 & 0.321 & 0.378 & 0.083 & 0.123 & 0.165 & 0.172\tabularnewline
Beta & 0.240 & 0.359 & 0.470 & 0.564 & 0.204 & 0.323 & 0.429 & 0.487 & 0.103 & 0.160 & 0.216 & 0.211\tabularnewline
PSO & 0.266 & 0.377 & 0.484 & 0.557 & 0.239 & 0.332 & 0.420 & 0.458 & 0.134 & 0.183 & 0.220 & 0.203\tabularnewline
NES & \textbf{0.297} & \textbf{0.426} & \textbf{0.515} & \textbf{0.584} & \textbf{0.262} & 0.378 & 0.463 & 0.458 & \textbf{0.141} & 0.199 & 0.223 & 0.185\tabularnewline
SPSA & 0.292 & 0.407 & 0.468 & 0.490 & 0.253 & \textbf{0.397} & \textbf{0.499} & \textbf{0.537} & 0.131 & \textbf{0.214} & \textbf{0.260} & \textbf{0.275}\tabularnewline
\midrule
	\rowcolor{lime!30}\multicolumn{13}{c}{Fashion-MNIST \qquad $N=k$}\tabularnewline
Rand & 0.204 & 0.289 & 0.346 & 0.302 & 0.146 & 0.181 & 0.186 & 0.124 & 0.053 & 0.062 & 0.049 & 0.021\tabularnewline
Beta & 0.237 & 0.342 & 0.372 & 0.275 & 0.183 & 0.236 & 0.218 & 0.106 & 0.072 & 0.079 & 0.058 & 0.020\tabularnewline
PSO & 0.252 & 0.342 & 0.388 & 0.284 & \textbf{0.198} & 0.240 & 0.219 & 0.081 & \textbf{0.080} & 0.082 & 0.046 & 0.013\tabularnewline
NES & \textbf{0.274} & \textbf{0.360} & 0.381 & 0.282 & \textbf{0.198} & 0.234 & 0.213 & 0.113 & 0.071 & 0.076 & 0.055 & 0.016\tabularnewline
SPSA & \textbf{0.274} & \textbf{0.360} & \textbf{0.412} & \textbf{0.427} & 0.188 & \textbf{0.251} & \textbf{0.287} & \textbf{0.298} & 0.067 & \textbf{0.086} & \textbf{0.091} & \textbf{0.095}\tabularnewline
\bottomrule
\end{tabular}
}
\vspace{-0.7em}
\caption{Black-box OA on Fashion-MNIST dataset. In the $N=\infty$ experiments,
	($\taun$, mR) are reported in each cell, while only $\taun$ is
	reported in the cells when $N$ equals $50$ or $k$.
	A larger $k$ and a smaller $N$ make the attack harder.
	}
\label{tab:black-fashion}
\ifcvpr
\end{table*}
\else
\end{table}
\fi


\vspace{-5pt}
\subsection{Black-Box Order Attack Experiments}
\label{sec:expblack}
\vspace{-2pt}

To simulate a real-world attack scenario, we convert the ranking
models trained for Sec.~\ref{sec:expwhite} into black-box versions,
which are subject to limitations discussed in Sec.~\ref{sec:black}.
Black-box OA experiments are conducted on these models.

\textcolor{\blue}{
To optimize the surrogate loss $\taun$, we adopt and compare several black-box
optimizers:
(1) Random Search (Rand), which independently samples
every dimension of $\vr$ from uniform distribution
$\mathcal{U}(-\varepsilon,+\varepsilon)$, then clips it to $\Omega_\vq$;
(2) Beta-Attack (Beta), a modification of $\mathcal{N}$-Attack~\citep{nattack}
that generates the $\vr$ from an iteratively-updated Beta
distribution (instead of Gaussian) per dimension;
(3) Particle Swarm Optimization (PSO)~\citep{stdPSO}, a classic meta-heuristic
black-box optimizer with an extra step that clips the adversarial perturbation to $\Omega_\vq$;
(4) NES~\citep{nes-atk,nes-algo}, which
performs PGD~\cite{madry} using estimated gradient;
(5) SPSA~\citep{spsa-atk,spsa-algo},
which can be interpreted as NES using a different sampling distribution.
}

\begin{table}[t]
\resizebox{1.0\columnwidth}{!}{%
\begin{tabular}{cc|ccccc|c}
	\toprule
	Dataset & $k$ & Rand & Beta & PSO & NES & SPSA & SRC Time\tabularnewline
 \midrule
\cellcolor{lime!15}{\small Fashion-MNIST} &  5 & 0.195 & 0.386 & 0.208 & 0.208 & 0.202 & 0.080 \tabularnewline
\cellcolor{lime!15}{\small Fashion-MNIST} & 10 & 0.206 & 0.404 & 0.223 & 0.214 & 0.213 & 0.087 \tabularnewline
\cellcolor{lime!15}{\small Fashion-MNIST} & 25 & 0.228 & 0.435 & 0.249 & 0.236 & 0.235 & 0.108 \tabularnewline
\hline
\cellcolor{cyan!15}SOP &  5 & 1.903 & 2.638 & 1.949 & 1.882 & 1.783 & 0.091\tabularnewline
\cellcolor{cyan!15}SOP & 10 & 1.923 & 2.720 & 1.961 & 1.954 & 1.836 & 0.095\tabularnewline
\cellcolor{cyan!15}SOP & 25 & 1.936 & 2.745 & 1.985 & 1.975 & 1.873 & 0.117\tabularnewline
\bottomrule
\end{tabular}
}
	\vspace{-0.7em}
	\caption{\textcolor{\blue}{Run time (second) per $(k,50)$-OA adversarial example for different black-box methods.
	$\varepsilon{=}\frac{4}{255}$ and $Q{=}10^3$.}}
\label{tab:time}
\end{table}


We first investigate the black-box $(5,\infty)$-OA, as shown in the upper part ($N{=}\infty$)
of Tab.~\ref{tab:black-fashion} and Tab.~\ref{tab:black-sop}.
In these cases, $\taun$ does not pose any semantics-preserving penalty since $N{=}\infty$,
\textcolor{\blue}{which is similar to white-box attack with $\xi{=}0$.}
With the Rand optimizer, $\taun$ can be optimized to $0.309$ with $\varepsilon{=}\frac{4}{255}$ on Fashion-MNIST.
As $\varepsilon$ increases, we obtain better $\taun$ results, and
larger mR values as an expected side-effect.

\ifcvpr
\begin{table*}[t]
\else
\begin{table}[t]
\fi
\centering
\resizebox{1.0\linewidth}{!}{
\setlength{\tabcolsep}{0.3em}%

\begin{tabular}{c|cccc|cccc|cccc}
	\toprule
	\multirow{2}{*}{\textbf{Algorithm}} & \multicolumn{4}{c|}{$k=5$} & \multicolumn{4}{c|}{$k=10$} & \multicolumn{4}{c}{$k=25$}\tabularnewline
\cline{2-13} \cline{3-13} \cline{4-13} \cline{5-13} \cline{6-13} \cline{7-13} \cline{8-13} \cline{9-13} \cline{10-13} \cline{11-13} \cline{12-13} \cline{13-13}
 & \multicolumn{1}{c|}{$\varepsilon=\frac{2}{255}$} & \multicolumn{1}{c|}{$\varepsilon=\frac{4}{255}$} & \multicolumn{1}{c|}{$\varepsilon=\frac{8}{255}$} & $\varepsilon=\frac{16}{255}$ & \multicolumn{1}{c|}{$\varepsilon=\frac{2}{255}$} & \multicolumn{1}{c|}{$\varepsilon=\frac{4}{255}$} & \multicolumn{1}{c|}{$\varepsilon=\frac{8}{255}$} & $\varepsilon=\frac{16}{255}$ & \multicolumn{1}{c|}{$\varepsilon=\frac{2}{255}$} & \multicolumn{1}{c|}{$\varepsilon=\frac{4}{255}$} & \multicolumn{1}{c|}{$\varepsilon=\frac{8}{255}$} & $\varepsilon=\frac{16}{255}$\tabularnewline
\hline
None & 0.0, 2.0 & 0.0, 2.0 & 0.0, 2.0 & 0.0, 2.0 & 0.0, 4.5 & 0.0, 4.5 & 0.0, 4.5 & 0.0, 4.5 & 0.0, 12.0 & 0.0, 12.0 & 0.0, 12.0 & 0.0, 12.0\tabularnewline
\midrule
	\rowcolor{cyan!10}\multicolumn{13}{c}{Stanford Online Product \qquad $N=\infty$}\tabularnewline
Rand & 0.187, 2.6 & 0.229, 8.5 & 0.253, 85.8 & 0.291, 649.7 & 0.167, 5.6 & 0.197, 13.2 & 0.208, 92.6 & 0.222, 716.4 & 0.093, 14.1 & 0.110, 27.6 & 0.125, 146.7 & 0.134, 903.7\tabularnewline
Beta & 0.192, 3.3 & 0.239, 15.3 & 0.265, 176.7 & 0.300, 1257.7 & 0.158, 6.2 & 0.186, 19.9 & 0.207, 139.0 & 0.219, 992.5 & 0.099, 15.5 & 0.119, 37.1 & 0.119, 206.5 & 0.132, 1208.5\tabularnewline
PSO & 0.122, 2.1 & 0.170, 3.0 & 0.208, 13.3 & 0.259, 121.4 & 0.135, 4.8 & 0.177, 6.5 & 0.206, 22.8 & 0.222, 166.5 & 0.104, 12.7 & 0.122, 16.7 & 0.137, 49.5 & 0.140, 264.2\tabularnewline
NES & \textbf{0.254, 3.4} & 0.283, 15.6 & \textbf{0.325, 163.0} & \textbf{0.368, 1278.7} & \textbf{0.312, 7.2} & \textbf{0.351, 26.3} & 0.339, 227.1 & 0.332, 1486.7 & \textbf{0.242, 18.0} & \textbf{0.259, 51.5} & 0.250, 324.1 & 0.225, 1790.8\tabularnewline
SPSA & 0.237, 3.5 & \textbf{0.284, 11.9} & 0.293, 75.2 & 0.318, 245.1 & 0.241, 7.8 & 0.325, 22.2 & \textbf{0.362, 112.7} & \textbf{0.383, 389.0} & 0.155, 18.1 & 0.229, 41.9 & \textbf{0.286, 185.6} & \textbf{0.306, 557.8}\tabularnewline
\midrule
	\rowcolor{cyan!20}\multicolumn{13}{c}{Stanford Online Product \qquad $N=50$}\tabularnewline
Rand & 0.180 & 0.216 & 0.190 & 0.126 & 0.163 & 0.166 & 0.119 & 0.055 & 0.092 & 0.055 & 0.016 & 0.003\tabularnewline
Beta & 0.181 & 0.233 & 0.204 & 0.119 & 0.153 & 0.168 & 0.116 & 0.054 & 0.084 & 0.057 & 0.021 & 0.003\tabularnewline
PSO & 0.122 & 0.173 & 0.183 & 0.153 & 0.135 & 0.164 & 0.137 & 0.081 & 0.093 & 0.083 & 0.042 & 0.011\tabularnewline
NES & \textbf{0.247} & 0.283 & 0.246 & 0.152 & \textbf{0.314} & 0.295 & 0.195 & 0.077 & \textbf{0.211} & \textbf{0.136} & 0.054 & 0.013\tabularnewline
SPSA & 0.241 & \textbf{0.287} & \textbf{0.297} & \textbf{0.303} & 0.233 & \textbf{0.298} & \textbf{0.298} & \textbf{0.292} & 0.125 & 0.130 & \textbf{0.114} & \textbf{0.103}\tabularnewline
\midrule
	\rowcolor{cyan!30}\multicolumn{13}{c}{Stanford Online Product \qquad $N=k$}\tabularnewline
Rand & 0.148 & 0.100 & 0.087 & 0.026 & 0.094 & 0.044 & 0.018 & 0.001 & 0.023 & 0.009 & 0.002 & 0.001\tabularnewline
Beta & 0.136 & 0.106 & 0.053 & 0.025 & 0.076 & 0.040 & 0.010 & 0.004 & 0.021 & 0.004 & 0.001 & 0.001\tabularnewline
PSO & 0.102 & 0.098 & 0.059 & 0.031 & 0.088 & 0.049 & 0.022 & 0.007 & 0.040 & 0.015 & 0.006 & 0.001\tabularnewline
NES & \textbf{0.185} & 0.139 & 0.076 & 0.030 & \textbf{0.173} & 0.097 & 0.036 & 0.008 & \textbf{0.071} & \textbf{0.027} & 0.007 & 0.005\tabularnewline
SPSA & 0.172 & \textbf{0.154} & \textbf{0.141} & \textbf{0.144} & 0.107 & \textbf{0.104} & \textbf{0.085} & \textbf{0.069} & 0.026 & 0.025 & \textbf{0.017} & \textbf{0.016}\tabularnewline
\bottomrule
\end{tabular}
}
\vspace{-0.7em}
\caption{Black-box OA on Stanford Online Product dataset. In the $N=\infty$ experiments,
	($\taun$, mR) are reported in each cell, while only $\taun$ is
	reported in the cells when $N$ equals $50$ or $k$.
	A larger $k$ and a smaller $N$ make the attack harder.
	}
\label{tab:black-sop}
\ifcvpr
\end{table*}
\else
\end{table}
\fi

\textcolor{\blue}{
Since all the queries of Rand are independent, one intuitive way
towards better performance is to leverage the historical query results and 
adjust the search distribution.
Following this, we modify $\mathcal{N}$-Attack~\citep{nattack} into
Beta-Attack, replacing its Gaussian distributions into Beta distributions,
from which each element of $\vr$ is independently generated.
%
%
All the Beta distribution parameters are initialized as $1$,
where Beta degenerates into Uniform distribution (Rand).
During optimization, the probability density functions are modified
according to $\taun$ results, in order to increase the expectation of
the next adversarial perturbation drawn from these distributions.
Results in Tab.~\ref{tab:black-fashion} suggest Beta's advantage
against Rand, but it also shows that the Beta distributions
are insufficient for modeling the perturbation $\vr$ for OA.}

According to the $(k,\infty)$-OA results in Tab.~\ref{tab:black-fashion}
and Tab.~\ref{tab:black-sop}, NES and SPSA outperform Rand, Beta and PSO.
This means black-box optimizers based on estimated gradient are still the most
effective and efficient for $(k,\infty)$-OA.
A larger $\varepsilon$ leads to a unanimous increase
in $\taun$ metric, and a side effect of worse (larger)
mR.
Predictably, when $N{=}50$ or $k$, the algorithms may confront a great penalty
due to the absence of the selected candidates from the top-$N$ visible range.

Further results of $(k,50)$-OA and $(k,k)$-OA confirm our speculation,
as shown in the middle ($N{=}50$) and bottom ($N{=}k$) parts of Tab.~\ref{tab:black-fashion} and
Tab.~\ref{tab:black-sop}.
With $N{=}50$ and a fixed $\varepsilon$, algorithms that result in a small mR (especially for those with $\text{mR}{<}\frac{50}{2}$)
also perform comparably as in $(k,\infty)$-OA.
Conversely, algorithms that lead to a large mR in $(k,\infty)$-OA
are greatly penalized in $(k,50)$-OA.
The results also manifest a special characteristic of OA, that $\taun$ peaks at a certain
small $\varepsilon$, and does not positively correlate with $\varepsilon$.
This is rather apparent in difficult settings such as $(k,k)$-OA
on SOP dataset.
%
%
In brief, the optimizers based on estimated gradients
still perform the best in $(k,50)$-OA and $(k,k)$-OA,
and an excessively large perturbation budget is not necessary.

%

\textcolor{\blue}{
All these experiments demonstrate the effectiveness of optimizing the
surrogate objective $\taun$ to conduct black-box OA.
As far as we know, optimizers based on gradient estimation
are the most reliable choices.
Next, we adopt SPSA and perform practical OA
in real-world applications.}

\setlength{\textfloatsep}{8pt}
\begin{table}[t]
\resizebox{1.0\columnwidth}{!}{%
	\setlength{\tabcolsep}{3pt}
\begin{tabular}{c|c|ccc|c|c|c|c|c}
\toprule
\rowcolor{gray!10}\textbf{Algorithm} & $\varepsilon$ & $k$ & $Q$ & $\mathsfit{T}$ & Mean $\taun$ & Stdev $\taun$ & Max $\taun$ & Min $\taun$ & Median $\taun$ \tabularnewline
\midrule
SPSA & 1/255 & 5 & 100 & 204 & 0.390 & 0.373 & 1.000 & -0.600 & 0.400 \tabularnewline

SPSA & 1/255 & 10 & 100 & 200 & 0.187 & 0.245 & 0.822 & -0.511 & 0.200 \tabularnewline

SPSA & 1/255 & 25 & 100 & 153 & 0.039 & 0.137 & 0.346 & -0.346 & 0.033 \tabularnewline
\bottomrule
\end{tabular}
}
\vspace{-0.7em}
\caption{\textcolor{\blue}{Quantitative $(k,50)$-OA Results on JD Snapshop.}}
\label{tab:jd}
\end{table}

\textbf{Time Complexity.} Although the complexity of Algo.~\ref{algo:taun}
is $\mathcal{O}(k^2)$, the actual run time of our
Rust\footnote{Rust Programming language. See https://www.rust-lang.org/.}-based
SRC implementation is short, as measured with Python cProfile with a Xeon 5118
CPU and a GTX1080Ti GPU.
As shown in Tab.~\ref{tab:time},
SRC calculation merely consumes $0.117$ seconds on average across
the five algorithms for an adversarial example on SOP with $k=25$.
The overall time consumption is dominated by sorting
and PyTorch~\cite{pytorch} model inference.
Predictably, in real-world attack scenarios,
It is highly likely that the time consumption bottleneck stems from
other factors irrelevant to our method, such as network delays and bandwidth, or the query-per-second limit set by the service provider.

\subsection{Practical Black-Box Order Attack Experiments}
\label{sec:jd}
 
``JD SnapShop''~\citep{jd} is an e-commerce platform
based on content-based image retrieval~\citep{cbir}.
Clients can upload query merchandise images via an {\tt HTTP}-protocol-based API,
and then obtain the top-$50$ similar products.
This exactly matches the setting of $(k,50)$-OA.
As the API specifies a file size limit ($\leqslant$ {\tt 3.75MB}), and
a minimum image resolution ($128{\times}128$),
we use the standard $(224{\times}224)$ size for the query.
%
We merely provide limited evaluations
because the API poses a hard limit of $500$ queries per day per user.

As noted in Sec.~\ref{sec:expblack}, the $\taun$
peaks with a certain small $\varepsilon$. Likewise, an empirical search for
$\varepsilon$ suggests that $1/255$ and $2/255$ are
the most suitable choices for OA against ``JD SnapShop''.
Any larger $\varepsilon$ value easily leads to the disappearance of $\sC$ from $\sX$.
This is meanwhile a preferable characteristic, as smaller perturbations
are less perceptible.

\textcolor{\blue}{
As shown in Fig.~\ref{fig:advorder}, we select the top-$5$ candidates as $\sC$,
and specify $\rvp{=}[1,5,4,3,2]$. Namely,
the expected \emph{relative order} among $\sC$ is
$\vc_1{\prec}\vc_5{\prec}\vc_4{\prec}\vc_3{\prec}\vc_2$.
By maximizing the $\taun$ with SPSA and 
$\varepsilon{=}\frac{1}{255}$,
we successfully convert the \emph{relative order} to the specified one
using $200$ times of queries.
This shows that performing OA by optimizing our proposed surrogate loss $\taun$
with a black-box optimizer is \emph{practical}.
Some limited quantitative results are presented in Tab.~\ref{tab:jd},
where $Q$ is further limited to $100$ in order to gather
more data, while a random permutation and a random query image from SOP
is used for each of the $\mathsfit{T}$ times of attacks.
}

\begin{table}
\resizebox{1.0\columnwidth}{!}{%
\setlength{\tabcolsep}{0.3em}
\renewcommand{\frac}[2]{#1/#2}
\begin{tabular}{c|c|rrr|c|c|c|c|c}
\toprule
\rowcolor{black!10}\textbf{Algorithm} & $\varepsilon$ & $k$ & $Q$ & $\mathsfit{T}$ & Mean $\tau_\mathsfit{S}$ & Stdev $\tau_\mathsfit{S}$ & Max $\tau_\mathsfit{S}$ & Min $\tau_\mathsfit{S}$ & Median $\tau_\mathsfit{S}$\tabularnewline
\midrule
SPSA & $\frac{8}{255}$ &  5 & 100 & 105 & 0.452 & 0.379 & 1.000 & -0.400 & 0.600\tabularnewline
SPSA & $\frac{8}{255}$ & 10 & 100 & 95  & 0.152 & 0.217 & 0.733 & -0.378 & 0.156\tabularnewline
SPSA & $\frac{8}{255}$ & 25 & 100 & 93  & 0.001 & 0.141 & 0.360 & -0.406 & 0.010\tabularnewline
\bottomrule
\end{tabular}
}
\vspace{-0.7em}
\caption{$(k,50)$-OA Results on Bing Visual Search API.}
\label{tab:bing}
\end{table}

We also conduct OA against Bing Visual Search API~\cite{bing}
following the same protocol. We find this API less sensitive to
weak (\emph{i.e.}, $\varepsilon{=}\frac{1}{255}$) perturbations
unlike Snapshop, possibly due to a different data pre-processing
pipeline. As shown in Tab.~\ref{tab:bing}, our OA is also
effective against this API.

In practice, the adversarial example may be slightly
changed by transformations such as the \emph{lossy} JPEG/PNG compression,
and resizing (as a pre-processing step),
which eventually leads to changes on the $\taun$ surface.
%
But a black-box optimizer should be robust to such influences.


%

\section{Conclusion}

Deep ranking systems inherited the adversarial vulnerabilities of deep
neural networks.
In this paper, the \emph{Order Attack} is proposed to tamper the \emph{relative order} among
selected candidates.
Multiple experimental evaluations of the white-box and black-box \emph{Order Attack}
illustrate their effectiveness, as well as
the deep ranking systems’ vulnerability in practice.
Ranking robustness and fairness with respect to \emph{Order Attack} may be the next
valuable direction to explore.
Code: \url{https://github.com/cdluminate/advorder}.

\textbf{Acknowledgement.}
%
This work was supported partly by National Key R\&D Program of China Grant
2018AAA0101400, NSFC Grants 62088102, 61976171, and XXXXXXXX, and Young Elite
Scientists Sponsorship Program by CAST Grant 2018QNRC001.

\clearpage
{\small
\bibliographystyle{ieee_fullname}
\bibliography{iccv}
}

\clearpage
\appendix
\section{Order Attack against ``JD SnapShop'' API}

\subsection{More Technical Details}

	We present additional technical details about the
	``JD SnapShop'' API~\citep{jd} and the
	\ifsupp
	Fig.~1 :
	\else
	Fig.~\ref{fig:advorder}:
	\fi

	\begin{enumerate}[nosep]
		\item Since the perturbation tensor contains negative values, it is normalized
			before being displayed in
			\ifsupp
			Fig.~1 :
			\else
			Fig.~\ref{fig:advorder}:
			\fi
			\begin{equation}\text{Normalize}(\vr) = 0.5 + 0.5 * \vr / \max\big(\text{abs}(\vr)\big).\end{equation}
		\item Since the selected candidates $\sC$ are [\textcolor{red}{A},
			\textcolor{orange}{B}, \textcolor{green}{C}, \textcolor{cyan}{D},
			\textcolor{blue}{E}], and the permutation vector is
			$\rvp=[1,5,4,3,2]$, the expected relative order among $\sC$ is
			$\sC_{p_1}\prec \sC_{p_2}\prec \sC_{p_3}\prec \sC_{p_4}\prec \sC_{p_5}$,
			\ie,
			\textcolor{red}{A} $\prec$
			\textcolor{blue}{E} $\prec$
			 \textcolor{cyan}{D} $\prec$
			\textcolor{green}{C} $\prec$
			\textcolor{orange}{B}.
		\item Each product corresponds to multiple images. Only the default
			product images specified by the service provider are displayed in the figure.
		\item The API in fact provides a similarity score
			for every candidate, which
			indeed can be leveraged for, \eg, better gradient
			estimation.
			However, the other practical ranking applications
			may not necessarily provide these similarity scores.
			Hence, we simply ignore such discriminative information to
			deliberately increase the difficulty of attack.
			The concrete ways to take advantage of known similarity scores
			are left for future works.
		\item
			\ifsupp
			In Fig.~1, the original similarity scores of candidates
			\else
			In Fig.~\ref{fig:advorder}, the original similarity scores of candidates
			\fi
			from A
			to E are {\small [
			\textcolor{red}{$0.7132$},
			\textcolor{orange}{$0.6336$},
			\textcolor{green}{$0.6079$},
			\textcolor{cyan}{$0.5726$},
			\textcolor{blue}{$0.5700$}]}.
			With our adversarial query, the scores
			of A to E are changed into {\small [
				\textcolor{red}{$0.6960$},
				\textcolor{orange}{$0.5724$},
				\textcolor{green}{$0.5763$},
				\textcolor{cyan}{$0.5827$},
				\textcolor{blue}{$0.5898$}]}.
		\item The API supports a ``{\tt topK}'' argument, which enables the clients to 
			change the visible range $N$.
			We leave it as the recommended default value $50$.
		\item
			\ifsupp
			From Fig.~1, we note some visually duplicated images
			\else
			From Fig.~\ref{fig:advorder}, we note some visually duplicated images
			\fi
			among the candidates.
			For instance, there are many other
			candidates similar to candidate \textcolor{blue}{E},
			due to reasons such as different sellers reusing the same product image.
			These images are not adjacent to each other in the ranking list,
			since the platform assigns them with different similarity scores.
			For instance, the $5$-th and $8$-th candidates
			\ifsupp
			in the first row of Fig.~1 are assigned with
			\else
			in the first row of Fig.~\ref{fig:advorder} are assigned with
			\fi
			similarity scores {\small $[0.5700, 0.5521]$}, while the $2$-nd, $7$-th,
			and $10$-th items in the second row are assigned with similarity scores
			{\small $[0.5898, 0.5728, 0.5536]$}.
			Whether the calculation of similarity scores involves 
			multiple cues (\eg, by aggregating the similarity scores of multiple
			product images, or using information from other modalities such as text),
			or even engineering tricks are unknown and beyond
			the scope of discussion.
		\item Users (with the free plan) are merely allowed to perform
			$500$ times of queries per day as a hard limit.
		\item The API documentation can be found at\\
			{\scriptsize\tt\url{https://aidoc.jd.com/image/snapshop.html}}.
		\item The SKU ID atop every candidate image can be used to browse
			the real product webpages on the ``JingDong'' shopping platform.
			The URL format is\\
			{\scriptsize\tt https://item.jd.com/<SKU-ID>.html}\\
			For example, the webpage for the product with SKU ID {\small\tt 72210617020}
			is located at\\
			{\scriptsize\url{https://item.jd.com/72210617020.html}}\\
			Note, due to irresistible reasons such as sellers withdrawing
			their product webpage and the ranking
			algorithm/database updates,
			some of the links may become invalid during the review process,
			and the ranking result for the same query may change.
		\item
			It consumes 200 times of queries to find the adversarial
			example presented in
			\ifsupp
			Fig.~1.
			\else
			Fig.~\ref{fig:advorder}.
			\fi
			This process takes
			around 170 seconds, mainly due our limited network condition.
	\end{enumerate}

	\subsection{Empirical Search for $\varepsilon$ on the API}

		According to the white-box and black-box OA experiments in the manuscript,
		we learn that the perturbation budget $\varepsilon$ affects the OA performance.
		And it meanwhile controls the adversarial perturbation imperceptibility.
		Thus, we search for a proper $\varepsilon$ for OA against ``JD SnapShop''.
		Due to the limitation that only $500$ times of queries per day are
		allowed for each user, we merely present some empirical and qualitative
		observation for different $\varepsilon$ settings.
		
		As shown in Tab.~\ref{tab:jdeps},
		we test the ``JD SnapShop'' API with adversarial query images using 
		the Rand algorithm with different $\varepsilon$ values.
		We conduct $50$ times of attack per value.
		Our qualitative observation is summarized in the table.

		\begin{table*}[t]
			\centering
			\resizebox{0.85\linewidth}{!}{%
			\begin{tabular}{rl}
				\toprule
				$\varepsilon$ & Empirical Qualitative Observation \\
				\midrule
$16/255$ & Almost any $\forall \vc \in \sC$ disappear from the top-$N$ candidates, resulting in $\taun\approx-1$.\\
$ 8/255$ & In most cases only $0\sim 1$ selected candidate remains within the top-$N$ result.\\
$ 4/255$ & In most cases only $1\sim 3$ selected candidates remain within the top-$N$ result.\\
$ 2/255$ & Nearly all $\vc \in \sC$ remain in top-$N$ with significant order change. (suitable for $\rvp$ with $p_1\neq 1$)\\
$ 1/255$ & Top-$1$ seldom moves. The rest part is slightly changed. (suitable for $\rvp$ with $p_1=1$)\\
 				\bottomrule
			\end{tabular}
				}
			\caption{Empirical \& Qualitative Parameter Search for $\varepsilon$ on ``JD SnapShop''.}
			\label{tab:jdeps}
		\end{table*}

		From the table, we find that $\varepsilon=1/255$ and $\varepsilon=2/255$
		are the most suitable choices for the $(5,50)$-OA against ``JD SnapShop''.
		This is meanwhile very preferable since such slight perturbations are
		imperceptible to human.
		As shown in Fig.~\ref{fig:imperceptible}, the $\varepsilon=1/255, 2/255, 4/255$
		adversarial perturbation can hardly be perceived, but the largest 
		perturbation (\ie, $\varepsilon=16/255$) used by \cite{i-fgsm} is
		relatively visible.
		%

		\begin{figure*}[t]
			\centering
			\begin{minipage}{0.16\textwidth}
				\centering
				\includegraphics[width=1.0\textwidth]{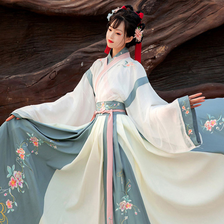}
				$\varepsilon=0$
			\end{minipage}
			\begin{minipage}{0.16\textwidth}
				\centering
				\includegraphics[width=1.0\textwidth]{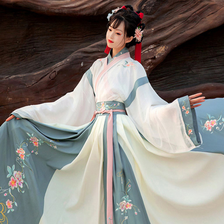}
				$\varepsilon=1/255$
			\end{minipage}
			\begin{minipage}{0.16\textwidth}
				\centering
				\includegraphics[width=1.0\textwidth]{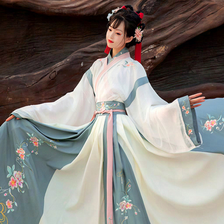}
				$\varepsilon=2/255$
			\end{minipage}
			\begin{minipage}{0.16\textwidth}
				\centering
				\includegraphics[width=1.0\textwidth]{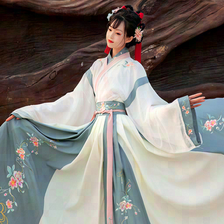}
				$\varepsilon=4/255$
			\end{minipage}
			\begin{minipage}{0.16\textwidth}
				\centering
				\includegraphics[width=1.0\textwidth]{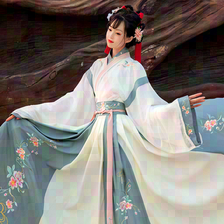}
				$\varepsilon=8/255$
			\end{minipage}
			\begin{minipage}{0.16\textwidth}
				\centering
				\includegraphics[width=1.0\textwidth]{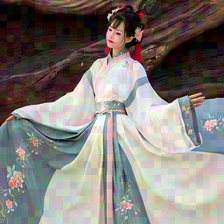}
				$\varepsilon=16/255$
			\end{minipage}
			\caption{Imperceptibility: Images perturbed under different perturbation budgets.
			A large perturbation with $\varepsilon=16/255$ is already relatively
			visible. However, in fact,
			$\varepsilon=1/255$ and $\varepsilon=2/255$ are more suitable choices
			for OA against ``JD SnapShop'', because they are least visible,
			and may lead to the best $\taun$.}
				\label{fig:imperceptible}
		\end{figure*}

	\subsection{More Showcases of OA against the API}

		\begin{itemize}[nosep]
			\item In showcase \#2 (Fig.\ref{fig:advorder2}), the original similarity
				scores of the top-$5$ candidates are {\small[
			\textcolor{red}{$0.7551$}, \textcolor{orange}{$0.6586$},
			\textcolor{green}{$0.6586$}, \textcolor{cyan}{$0.6533$},
				\textcolor{blue}{$6507$}]}.
			They are changed into {\small[
			\textcolor{red}{$0.6748$}, \textcolor{orange}{$0.6723$},
			\textcolor{green}{$0.6723$}, \textcolor{cyan}{$0.6921$},
				\textcolor{blue}{$0.6609$}]} with the perturbation.

	\item In showcase \#3 (Fig.\ref{fig:advorder3}), the original top-$5$
		candidate similarity {\small[
			\textcolor{red}{$0.8522$}, \textcolor{orange}{$0.8333$},
			\textcolor{green}{$0.8341$}, \textcolor{cyan}{$0.7659$},
				\textcolor{blue}{$0.7159$}]} is changed into
			{\small[
			\textcolor{red}{$0.8792$}, \textcolor{orange}{$0.7928$},
			\textcolor{green}{$0.8470$}, \textcolor{cyan}{$0.7958$},
				\textcolor{blue}{$0.7130$}]}.

	\item Fig.~\ref{fig:miss1} shows two examples where one selected
		candidate disappear from top-$N$ result with our adversarial
		query. In the ``white shoes'' case, the top-$5$ candidate similarity scores
				are changed from {\small[
			\textcolor{red}{$0.8659$}, \textcolor{orange}{$0.8653$},
			\textcolor{green}{$0.8648$}, \textcolor{cyan}{$0.8619$},
				\textcolor{blue}{$0.8603$}]} to
			{\small[
			\textcolor{red}{N/A},    \textcolor{orange}{$0.8689$},
			\textcolor{green}{$0.8603$}, \textcolor{cyan}{$0.8640$},
				\textcolor{blue}{$0.8641$}]}.
		In the ``vase'' case, the top-$5$ candidate similarity scores
				are changed from {\small[
			\textcolor{red}{$0.9416$}, \textcolor{orange}{$0.9370$},
			\textcolor{green}{$0.9360$}, \textcolor{cyan}{$0.9350$},
				\textcolor{blue}{$0.9349$}]} to
			{\small[
			\textcolor{red}{N/A},    \textcolor{orange}{$0.9338$},
			\textcolor{green}{$0.9427$}, \textcolor{cyan}{$0.9392$},
				\textcolor{blue}{$0.9361$}]}.

	\item Fig.~\ref{fig:longtail} shows some long-tail queries on which
		our OA will not take effect, because a large portion of the top-ranked
		candidates have completely the same similarity scores.
		In the 1st row, \ie, results for a ``Machine Learning'' textbook query, the similarity
		scores of the $5$-th to $7$-th candidates are $0.9242$.
		The score of the $9$-th
		to $11$-th candidates are $0.9240$. That of the $22$-th to $50$-th are the same $0.9202525$.
		In the ``Deep Learning'' textbook case (2nd row), the similarity
		scores of the $27$-th to $50$-th candidates are $0.9288423$.
		In the ``RTX 3090 GPU'' case (3rd row), the similarity scores
		of the $7$-th to $9$-th candidate are unexceptionally $0.5581$.
		OA cannot change the \emph{relative order} among those candidates
		with the same similarity scores.

		\end{itemize}

	\section{Visualizing Black-Box OA on Fashion-MNIST \& Stanford Online Products}

	We present some visualizations of the black-box OA
	on the Fashion-MNIST dataset and the SOP dataset, as shown in
	(Fig.~\ref{fig:fb1}, Fig.~\ref{fig:fb2}, Fig.~\ref{fig:fb3}) and
	(Fig.~\ref{fig:sb1}, Fig.~\ref{fig:sb2}, Fig.~\ref{fig:sb3}), respectively.
	All the adversarial perturbations are found under $\varepsilon=4/255$
	and $N=50$.

	All these figures are picture matrices of size $(4, 2+2k)$.
	In particular, pictures at location $(1,3)$, $(1,5)$ and $(1,7)$ are
	the original query, the perturbation
	and the perturbed query image, respectively.
	The 2nd row in each figure is the original query and the corresponding
	ranking list (truncated to the top-$2k$ results).
	The 3rd row in each figure is the permuted top-$k$ candidates.
	The 4th row in each figure is the adversarial query and the corresponding
	ranking list (also truncated to the top-$2k$ results).
	Every picture is annotated with its \emph{ID} in the dataset
	and its \emph{label} for classification.



\section{Additional Information for White-Box OA}

\subsection{Example case where an ideal $\tilde{q}$ is impossible}

As discussed in Sec.~\ref{sec:white}, an ideal adversarial example
that fully satisfy the
desired relative order does not always exist. As an example,
let $\vv_q, \vv_1, \vv_2, \vv_3$ be the embeddings of $\tilde{\vq}, \vc_1, \vc_2, \vc_3$
mentioned in the last paragraph of Sec.~\ref{sec:white}, respectively.
Assume the embeddings are non-zero vectors, $\vv_1{=}\vv_2{-}\vo$, $\vv_3{=}\vv_2{+}\vo$,
where vector $\vo{\ne}0$.
Then a specified permutation $\vc_1{\prec}\vc_3{\prec}\vc_2$, namely
$f(\tilde{\vq},\vc_1){<}f(\tilde{\vq},\vc_3){<}f(\tilde{\vq},\vc_2)$
requires $-(\vv_q^T\vo-\vv_2^T\vo)>\vv_q^T\vo-\vv_2^T\vo$
and $\vv_q^T\vo-\vv_2^T\vo>0$, which are contradictory to each other.
Thus, there is no satisfactory $\vv_q$.

	\subsection{More Results on Ablation of $L_\text{QA+}$}

\begin{table}[h]
\centering
\resizebox{1.0\columnwidth}{!}{
\setlength{\tabcolsep}{0.3em}%
\begin{tabular}{c|ccccc|ccccc|ccccc}
\toprule
 & \multicolumn{5}{c|}{$k=5$} & \multicolumn{5}{c|}{$k=10$} & \multicolumn{5}{c}{$k=25$}\tabularnewline
\hline
$\varepsilon$ & 0 & $\frac{2}{255}$ & $\frac{4}{255}$ & $\frac{8}{255}$ & $\frac{16}{255}$ & 0 & $\frac{2}{255}$ & $\frac{4}{255}$ & $\frac{8}{255}$ & $\frac{16}{255}$ & 0 & $\frac{2}{255}$ & $\frac{4}{255}$ & $\frac{8}{255}$ & $\frac{16}{255}$\tabularnewline
\midrule
\multicolumn{16}{c}{\cellcolor{lime!10}Fashion-MNIST \qquad $\xi=0$}\tabularnewline
$\taun$ & 0.000 & 0.336 & 0.561 & 0.777 & 0.892 & 0.000 & 0.203 & 0.325 & 0.438 & 0.507 & 0.000 & 0.077 & 0.131 & 0.170 & 0.189\tabularnewline
mR & 2.0 & 5.5 & 27.2 & 52.7 & 75.6 & 4.5 & 8.0 & 17.3 & 40.4 & 63.4 & 12.0 & 16.4 & 19.2 & 22.8 & 25.3\tabularnewline
\midrule
\multicolumn{16}{c}{\cellcolor{cyan!10}Stanford Online Products \qquad $\xi=0$}\tabularnewline
$\taun$ & 0.000 & 0.932 & 0.970 & 0.975 & 0.975 & 0.000 & 0.632 & 0.760 & 0.823 & 0.832 & 0.000 & 0.455 & 0.581 & 0.646 & 0.659\tabularnewline
mR & 2.0 & 973.9 & 1780.1 & 2325.5 & 2421.9 & 4.5 & 1222.7 & 3510.2 & 5518.6 & 6021.4 & 12.0 & 960.4 & 2199.2 & 3321.4 & 3446.3\tabularnewline
\bottomrule
\end{tabular}
}
\caption{More Results of White-Box OA with $\xi=0$ (\ie, without the $L_\text{QA+}$ term). These results are supplementary
	to Tab.~2 in the manuscript.}
\label{tab:whitenosp}
\end{table}

	In order to make sure the selected candidates $\sC$ will not disappear
	from the top-$N$ results during the OA process, a semantics-preserving
	term $L_\text{QA+}$ is introduced to maintain the \emph{absolute ranks}
	of $\sC$.
	Tab.~2 in the manuscript presents two ablation experimental results of $L_\text{QA+}$
	for the white-box $(5,\infty)$-OA with $\varepsilon=4/255$.
	In this subsection, we provide the full ablation experiments of $L_\text{QA+}$
	in all parameter settings, as shown in Tab.~\ref{tab:whitenosp}.
	After
	removing the $L_\text{QA+}$ term from the loss function (\ie, setting $\xi=0$),
	the white-box OA can achieve a better $\taun$, meanwhile a worse mR.
	When comparing it with Tab.~1 in the manuscript, we find that
	(1) the semantics-preserving term $L_\text{QA+}$ is
	effective for keeping the selected $C$ within top-$N$ results;
	(2) $L_\text{QA+}$ will increase the optimization difficulty, so there will
	be a trade off between $L_\text{ReO}$ and $L_\text{QA+}$.
	\ifsupp
	These results support our analysis and discussion in Sec.~4.1.
	\else
	These results support our analysis and discussion in Sec.~\ref{sec:expwhite}.
	\fi

	\subsection{Transferability}

Some adversarial examples targeted at ranking models have been found
transferable~\cite{advrank}.
Following \cite{advrank},
we train Lenet and ResNet18 models on Fashion-MNIST besides the C2F1 model
used in the manuscript. Each of them is trained with two different
parameter initializations (annotated with \#1 and \#2). Then we conduct
transferability-based attack
using a white-box surrogate model with $k=5$ and $\varepsilon=\frac{4}{255}$,
as shown in Tab.~\ref{tab:transfer}.

\begin{table}[t]
\noindent\resizebox{1.0\columnwidth}{!}{%
	\setlength{\tabcolsep}{3pt}
\begin{tabular}{c|cc|cc|cc}
	\toprule
\textbf{From \textbackslash{} To} & Lenet \#1 & Lenet \#2 & C2F1 \#1 & C2F1 \#2 & ResNet18 \#1 & ResNet18 \#2\tabularnewline
\midrule
Lenet \#1 & \underline{0.377} & -0.003 & 0.013 & -0.014 & 0.003 & 0.010\tabularnewline
C2F1 \#1 & 0.016 & 0.003 & \underline{0.412} & 0.005 & 0.020 & 0.008\tabularnewline
ResNet18 \#1 & -0.001 & -0.011 & -0.006 & 0.020 & \underline{0.268} & 0.016\tabularnewline
\bottomrule
\end{tabular}
}
	\caption{OA Example Transferability Experiment.}
	\label{tab:transfer}
\end{table}

According to the table, OA adversarial example does not exhibit
transferability over different
architectures or different parameter initializations.
We believe these models learned distinct embedding spaces,
across which enforcing a specific fixed ordering is particularly
difficult without prior knowledge of the model being attacked.
%
As transferability is not a key point of the manuscript,
we only discuss it in this appendix.

	\subsection{Intra-Class Variance \& OA Difficulty}

	In the last paragraph of Sec.~4.1 (White-Box Order Attack Experiments),
	it is claimed that,
	\begin{quote}\it
		The intra-class variance of
the relatively simple Fashion-MNIST dataset is smaller than
that of SOP, which means samples of the same class are
densely clustered in the embedding space. As each update
could change the $\sX$ more drastically, it is more difficult to
adjust the query embedding position with a fixed PGD step
for a higher $\taun$ without sacrificing the mR value.
	\end{quote}
	To better illustrate the idea, a diagram is present in Fig.~\ref{fig:mnistmR}.
	Given a trained model, and a fixed PGD step size as $1/255$,
	the query image is projected near a dense embedding cluster in case I, while
	near a less dense embedding cluster in case II.
	Then the query image is updated with a fixed step size in both case I and case II,
	resulting in similar position change in the embedding space.
	At the same time the ranking list for the updated query will change as well.
	However, the ranking list in case I changes
	much more dramatically than that in case II.
	As a result, there is a higher chance for the ranks of the selected $\sC$
	to significantly change in case I, leading to a high value of mR.
	Besides, $1/255$ is already the smallest appropriate choice for the PGD
	update step size, as in practice the adversarial examples will be quantized
	into the range $[0,255]$.
	Namely, the position of the query embedding cannot be modified in a
	finer granularity given such limitation.
	Since a dataset with small intra-class variance tend to be densely clustered
	in the embedding space, we speculate that 
	optimizing $L_\text{OA}$ on a dataset with small intra-class variance
	(\eg, Fashion-MNIST) is difficult, especially in terms of maintaining a low mR.

	\begin{figure}[h]
		\includegraphics[width=1.0\columnwidth]{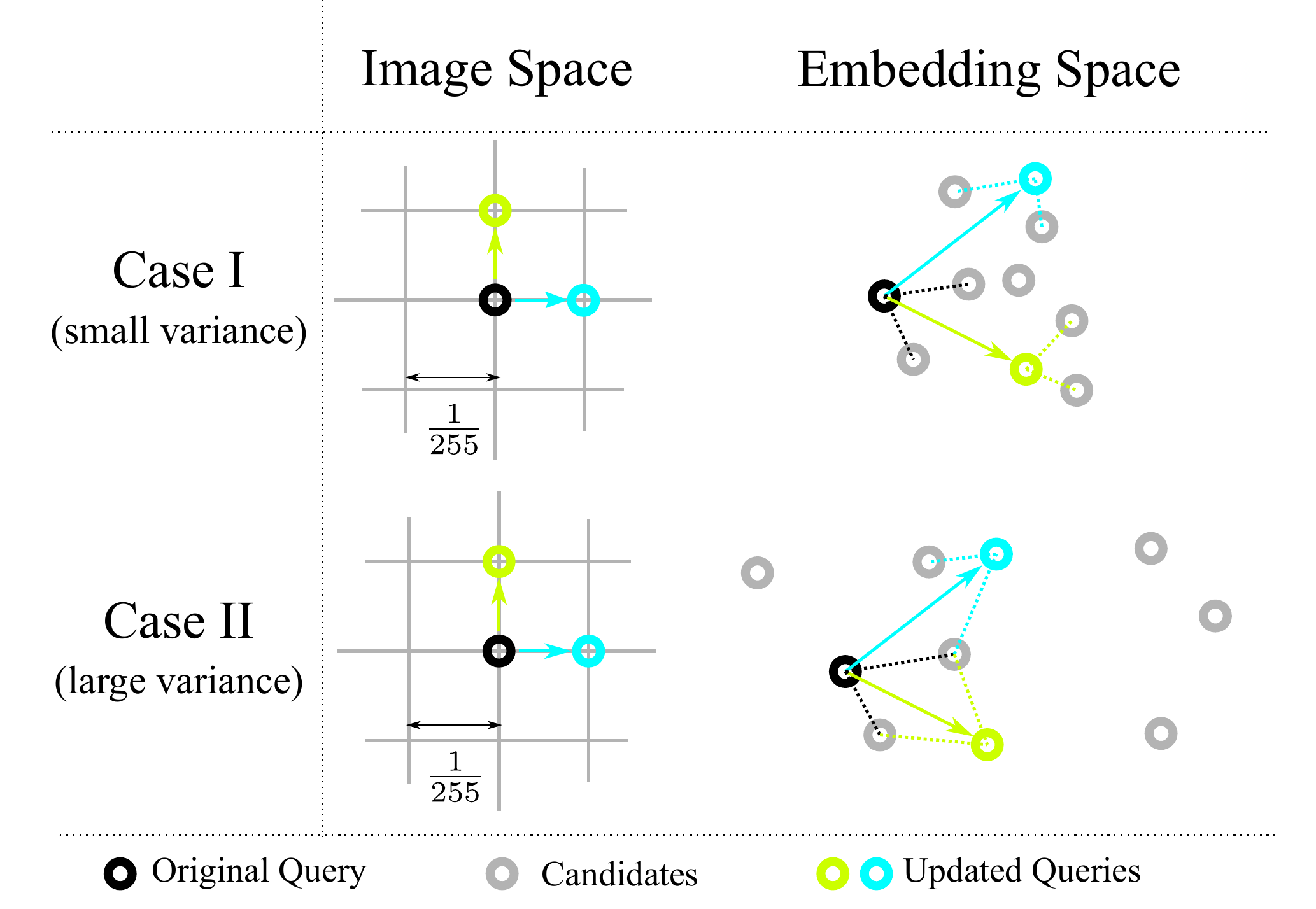}
		\caption{Intra-Class Variance \& OA Difficulty. Updated queries
		are linked to two closest candidates with doted lines.}
		\label{fig:mnistmR}
	\end{figure}

\begin{table*}[t]
	\centering
	\resizebox{1.0\linewidth}{!}{%
\begin{tabular}{c|cccc|cccc|cccc}
\toprule
	\multirow{2}{*}{\textbf{Algorithm}} & \multicolumn{4}{c|}{\cellcolor{cyan!10}$N=\infty$ ~~ $k=5$} & \multicolumn{4}{c|}{\cellcolor{cyan!20}$N=50$ ~~ $k=5$} & \multicolumn{4}{c}{\cellcolor{cyan!30}$N=5$ ~~ $k=5$}\tabularnewline
\cline{2-13} \cline{3-13} \cline{4-13} \cline{5-13} \cline{6-13} \cline{7-13} \cline{8-13} \cline{9-13} \cline{10-13} \cline{11-13} \cline{12-13} \cline{13-13}
 & \multicolumn{1}{c|}{$\varepsilon=\frac{2}{255}$} & \multicolumn{1}{c|}{$\frac{4}{255}$} & \multicolumn{1}{c|}{$\frac{8}{255}$} & $\frac{16}{255}$ & \multicolumn{1}{c|}{$\frac{2}{255}$} & \multicolumn{1}{c|}{$\frac{4}{255}$} & \multicolumn{1}{c|}{$\frac{8}{255}$} & $\frac{16}{255}$ & \multicolumn{1}{c|}{$\frac{2}{255}$} & \multicolumn{1}{c|}{$\frac{4}{255}$} & \multicolumn{1}{c}{$\frac{8}{255}$} & $\frac{16}{255}$\tabularnewline
\midrule
None & 0.0, 2.0 & 0.0, 2.0 & 0.0, 2.0 & 0.0, 2.0 & 0.0 & 0.0 & 0.0 & 0.0 & 0.0 & 0.0 & 0.0 & 0.0\tabularnewline
\rowcolor{gray!20}Rand (w/o DR) & 0.106, 2.1 & 0.151, 3.1 & 0.190, 13.1 & 0.224, 117.5 & 0.101 & 0.139 & 0.167 & 0.148 & 0.076 & 0.086 & 0.058 & 0.026\tabularnewline
Rand (w/ DR) & 0.187, 2.6 & 0.229, 8.5 & 0.253, 85.8 & 0.291, 649.7 & 0.180 & 0.216 & 0.190 & 0.126 & 0.148 & 0.100 & 0.087 & 0.026\tabularnewline
\rowcolor{gray!20}Beta (w/o DR) & 0.120, 2.2 & 0.158, 3.8 & 0.199, 21.8 & 0.231, 205.7 & 0.115 & 0.164 & 0.173 & 0.141 & 0.097 & 0.096 & 0.060 & 0.035\tabularnewline
Beta (w/ DR) & 0.192, 3.3 & 0.239, 15.3 & 0.265, 176.7 & 0.300, 1257.7 & 0.181 & 0.233 & 0.204 & 0.119 & 0.136 & 0.106 & 0.053 & 0.025\tabularnewline
\rowcolor{gray!20}PSO (w/o DR) & 0.128, 2.1 & 0.174, 3.1 & 0.219, 13.0 & 0.259, 122.0 & 0.133 & 0.175 & 0.199 & 0.155 & 0.097 & 0.095 & 0.060 & 0.036\tabularnewline
PSO (w/ DR) & 0.122, 2.1 & 0.170, 3.0 & 0.208, 13.3 & 0.259, 121.4 & 0.122 & 0.173 & 0.183 & 0.153 & 0.102 & 0.098 & 0.059 & 0.031\tabularnewline
\rowcolor{gray!20}NES (w/o DR) & 0.139, 2.3 & 0.192, 4.8 & 0.244, 31.9 & 0.266, 300.0 & 0.128 & 0.192 & 0.208 & 0.166 & 0.108 & 0.102 & 0.079 & 0.039\tabularnewline
NES (w/ DR) & 0.254, 3.4 & 0.283, 15.6 & 0.325, 163.0 & 0.368, 1278.7 & 0.247 & 0.283 & 0.246 & 0.152 & 0.185 & 0.139 & 0.076 & 0.030\tabularnewline
\rowcolor{gray!20}SPSA (w/o DR) & 0.135, 2.4 & 0.171, 3.9 & 0.209, 15.9 & 0.226, 45.2 & 0.140 & 0.176 & 0.205 & 0.223 & 0.108 & 0.110 & 0.146 & 0.143\tabularnewline
SPSA (w/ DR) & 0.237, 3.5 & 0.284, 11.9 & 0.293, 75.2 & 0.318, 245.1 & 0.241 & 0.287 & 0.297 & 0.303 & 0.172 & 0.154 & 0.141 & 0.144\tabularnewline
\bottomrule
\end{tabular}}
\caption{Ablation study of the search space Dimension Reduction (DR) trick for black-box OA with SOP dataset.}
\label{tab:nodr}
\end{table*}

\section{Additional Information for Black-Box OA}

	\subsection{Average query number for successful attack}

\begin{figure}[h]
\includegraphics[width=0.49\columnwidth]{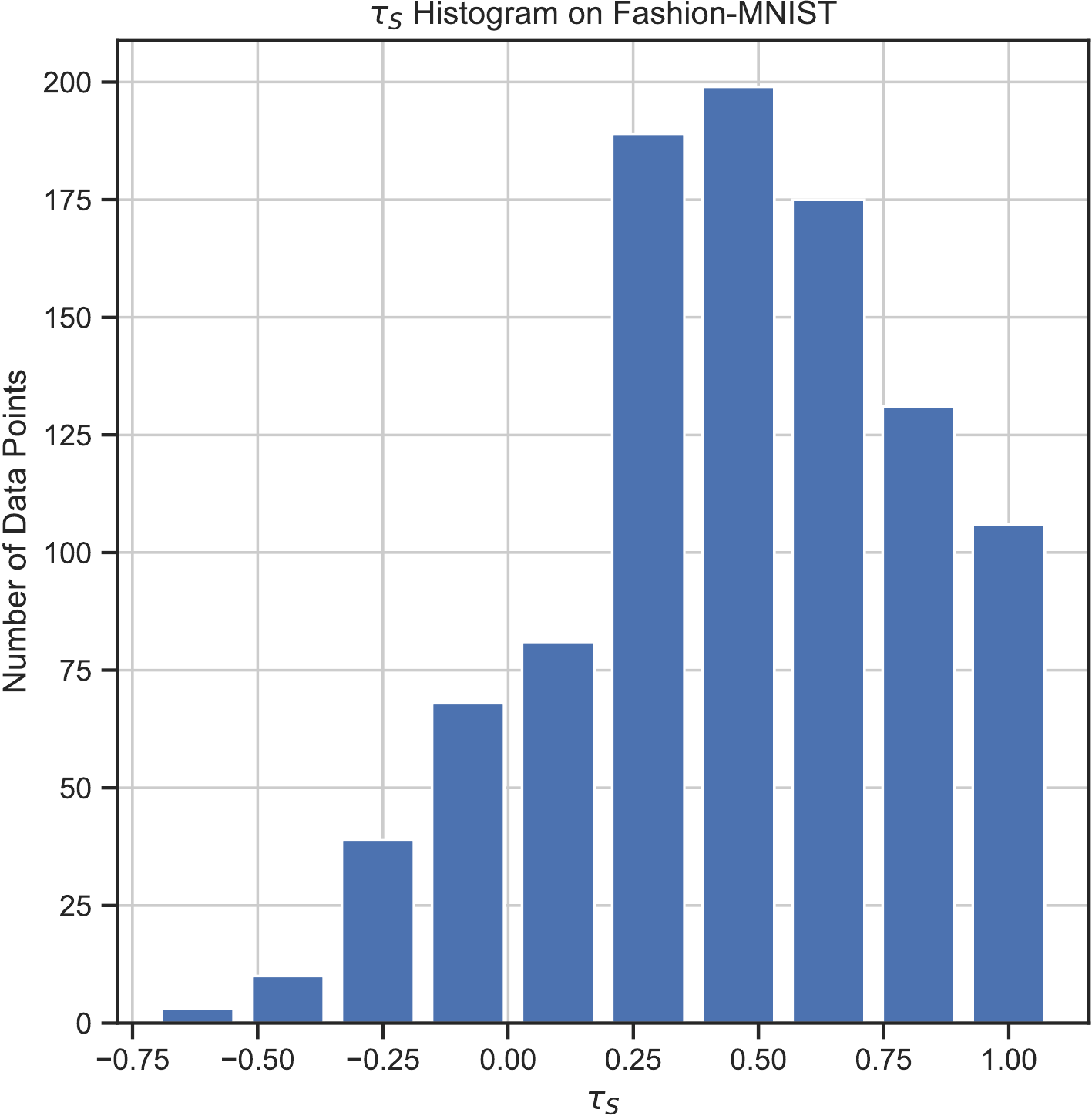}%
\includegraphics[width=0.49\columnwidth]{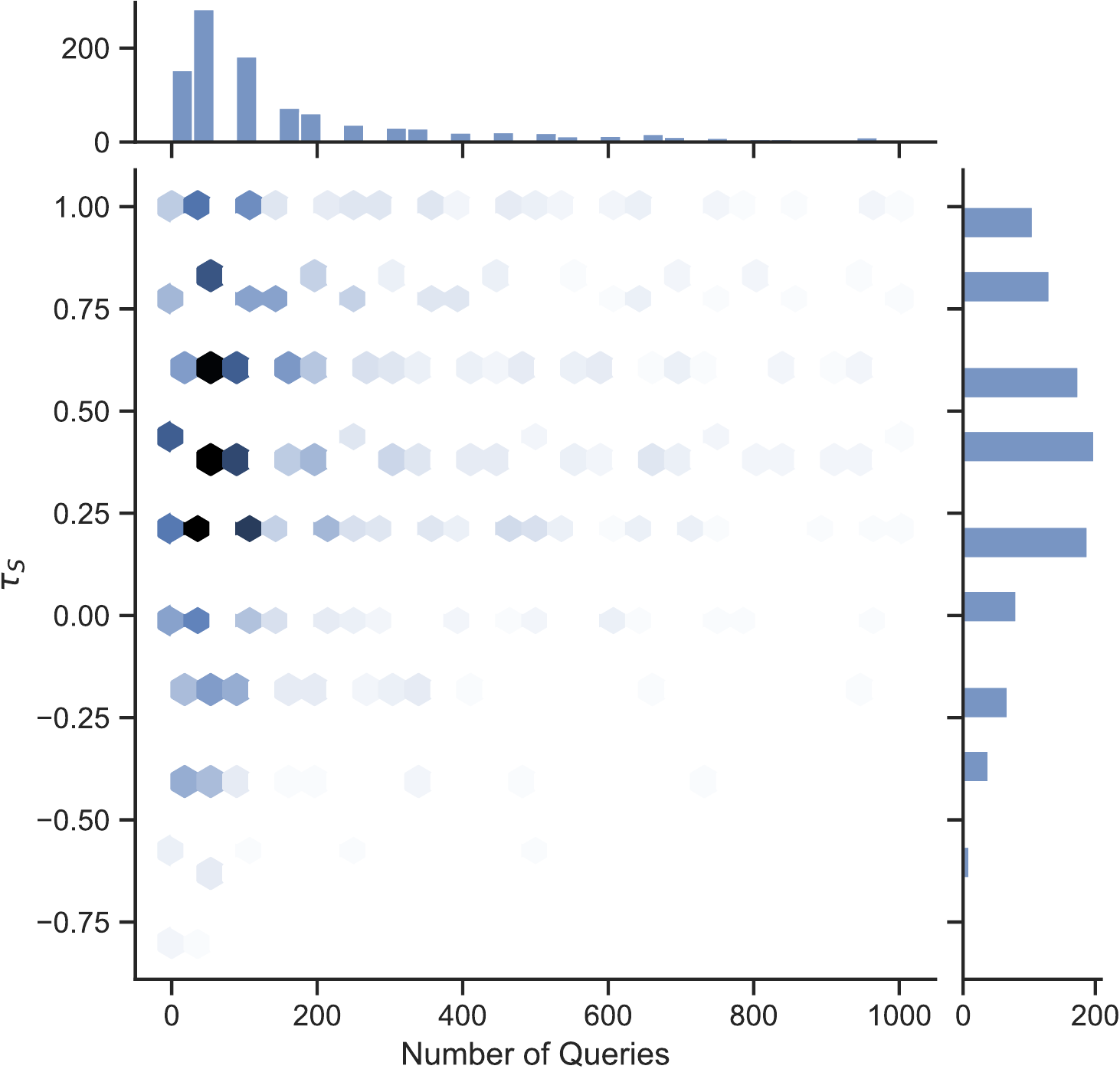}
	\caption{Histogram of $\taun$ (left) and hexbin plot with marginal
	histogram for $\taun$ and its respective query number (right) of
	$(5,50)$-OA on Fashion-MNIST.}
	\label{fig:fa-joint}
\end{figure}

	As shown in Fig.~\ref{fig:fa-joint},
we plot the histogram (left) of $\tau_\mathsfit{S}$ for $1000$ trials of $(5,50)$-OA
with SPSA ($\varepsilon=4/255$, $Q=10^3$) on Fashion-MNIST, as well as a
hexbin plot with marginal histogram for $\tau_\mathsfit{S}$ and its respective query number (right),
where $106$ trials reach $\tau_\mathsfit{S}=1.0$ with $226.4$ queries on average.

\begin{figure}[h]
\includegraphics[width=0.49\columnwidth]{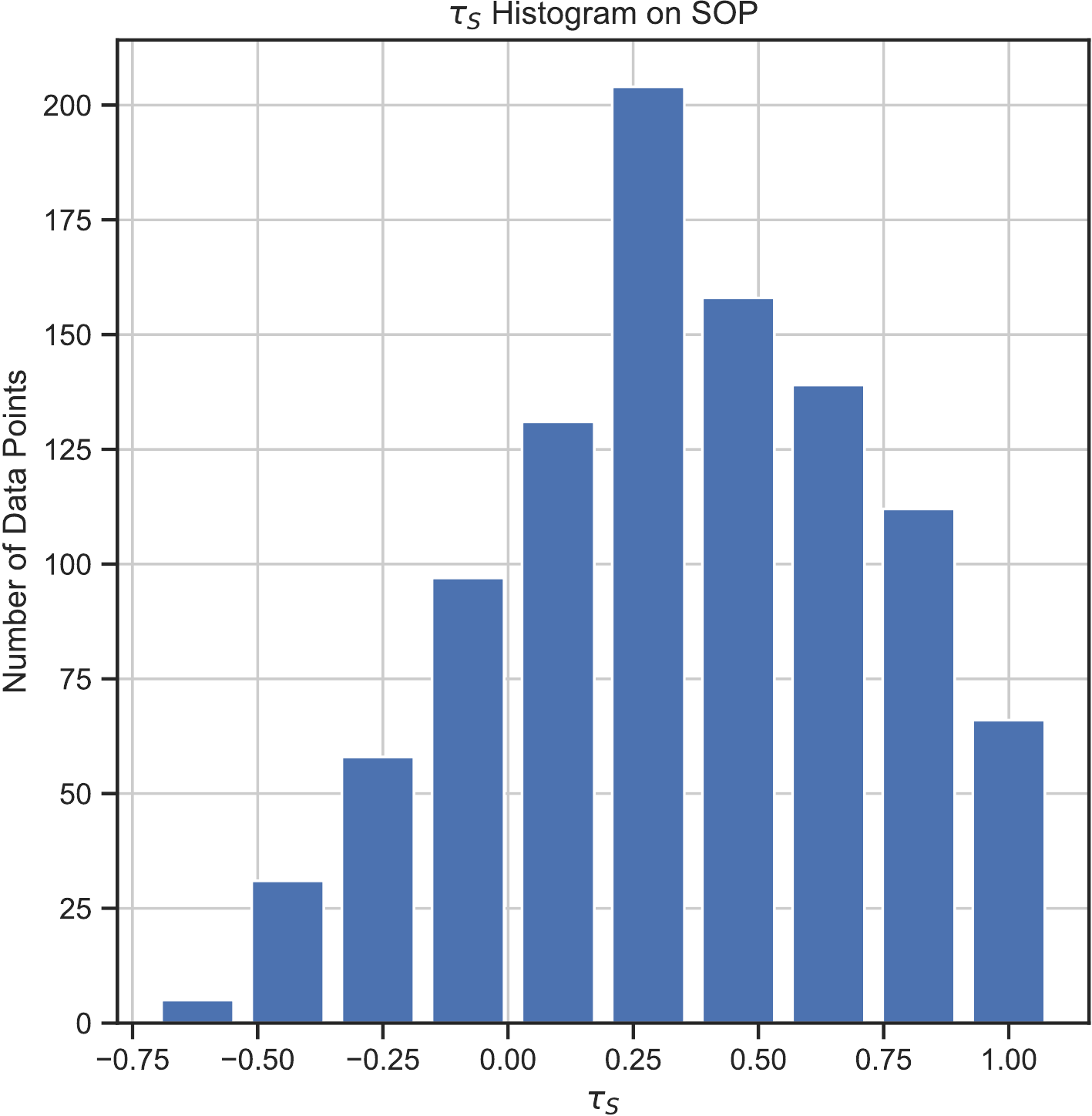}%
\includegraphics[width=0.49\columnwidth]{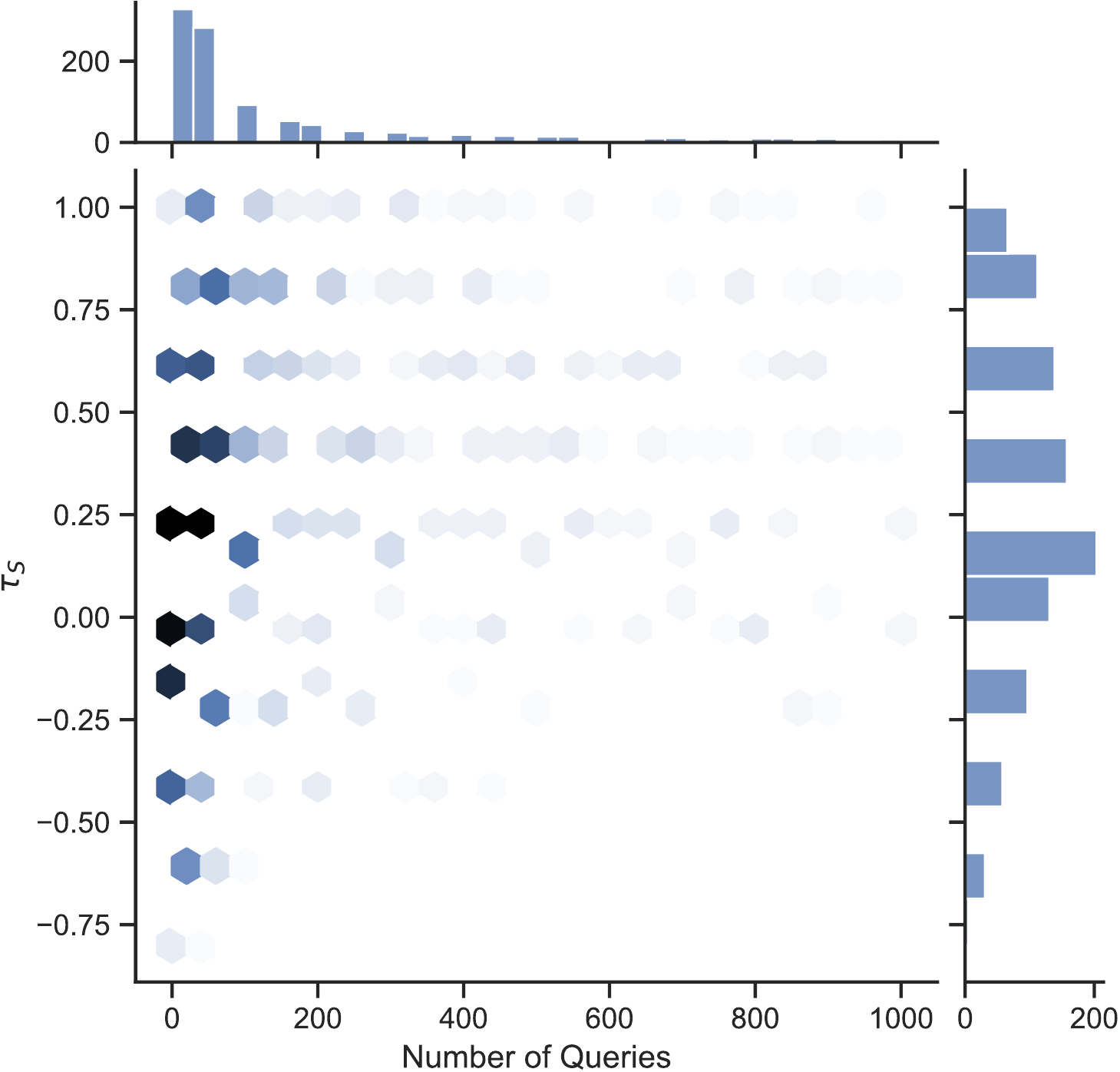}
	\caption{Histogram of $\taun$ (left) and hexbin plot with marginal
	histogram for $\taun$ and its respective query number (right) of
	$(5,50)$-OA on SOP.}
	\label{fig:sop-joint}
\end{figure}

Following the same setting, the plots with $1000$ trials of $(5,50)$-OA on SOP
is also available in Fig.~\ref{fig:sop-joint},
where $66$ trials reach $\tau_\mathsfit{S}=1.0$ with $213.6$ queries on average.

	\subsection{Relation between Query Budget $Q$ and $\taun$}

	Predictably, the query budget $Q$ could significantly impact
	the $\taun$ as it directly limits the max number of iterations
	for any given black-box optimizer.
	To study such impact, we conduct experiments on the Fashion-MNIST
	dataset, as shown in Tab.~\ref{tab:querybudget}.
	It is clear that the performance of all black-box
	optimizers become better with the query budget $Q$ increasing, but
	will eventually plateau.
	Even with an extremely limited query budget $Q=10^2$, the methods
	based on estimated gradients remain to be the most effective ones.


\begin{table}[h]
\centering
\resizebox{1.0\columnwidth}{!}{
\setlength{\tabcolsep}{0.6em}%
\begin{tabular}{c|ccccc}
	\toprule
	\multirow{2}{*}{\textbf{Algorithm}} & \multicolumn{5}{c}{\cellcolor{lime!10} Fashion-MNIST \qquad $N=\infty$, $k=5$, $\varepsilon=\frac{4}{255}$}\tabularnewline
\cline{2-6} \cline{3-6} \cline{4-6} \cline{5-6} \cline{6-6} 
 & \multicolumn{1}{c|}{$Q=10^{2}$} & \multicolumn{1}{c|}{$5\times10^{2}$} & \multicolumn{1}{c|}{{*}$10^{3}$} & \multicolumn{1}{c|}{$5\times10^{3}$} & $10^{4}$\tabularnewline
 \midrule
Rand & 0.233, 2.2 & 0.291, 2.2 & 0.309, 2.3 & 0.318, 2.2 & 0.320, 2.2\tabularnewline
Beta & 0.249, 2.2 & 0.313, 2.4 & 0.360, 2.6 & 0.368, 2.6 & 0.382, 2.4\tabularnewline
PSO & 0.280, 2.6 & 0.341, 2.4 & 0.381, 2.3 & 0.382, 2.4 & 0.385, 2.4\tabularnewline
NES & 0.309, 2.6 & 0.380, 2.9 & 0.416, 3.1 & 0.431, 2.9 & 0.438, 2.9\tabularnewline
SPSA & 0.292, 2.6 & 0.365, 2.8 & 0.407, 3.2 & 0.421, 2.9 & 0.433, 2.8\tabularnewline
\bottomrule
\end{tabular}
}
\caption{$\taun$ with different query budget $Q$.}
\label{tab:querybudget}
\end{table}


 	\subsection{Ablation of Search Space Dimension Reduction}
	\label{sec:dr}

	In this subsection, we study the effectiveness of the dimension reduction trick,
	which has been widely adopted in the literature~\citep{benchmarking,zoo,bayesian,qeba}.
	As shown in Tab.~\ref{tab:nodr}, all black-box optimizers benefit from
	this trick except for PSO, as illustrated by the performance gains.
	We leave the analysis on the special characteristics of PSO for future work.



%
\subsection{Random Initialization}
Some black-box classification attacks such as \cite{boundaryattack}
initializes the adversarial perturbation as a random vector.
However, we note that the adversarial perturbation for OA should be always
initialized as a zero vector, because random initialization is harmful.
For white-box OA, a random initial perturbation may dramatically change the
query semantics and push the query embedding off its original position 
by a large margin~\citep{advrank}.
Thus, the expectation of the initial mR will be higher,
hence an avoidable query semantics-preserving penalty will be triggered.
Then, the optimizer will have to 
``pull'' the adversarial query back
near its original location in the embedding space,
during which it may even stuck at a local optimum.
For black-box OA, the random initialization could be harmful for
the methods based on estimated gradient, such as NES and SPSA,
because the adversarial query may directly lie
on a ``flat'' area of the $\taun$ surface,
where all the selected candidates disappear from the top-$N$
and all the neighboring samples lead to $\taun=-1$ as well.
The estimated gradient will be invalid,
hence OA will fail.
In contrast, zero vector initialization could largely
mitigate such difficulties. 
%

\section{Defense Against OA}

According to \cite{advrank} which presents a defense method for deep ranking, our proposed
OA also requires the query embedding to be moved to a proper position.
Thus, the defense~\cite{advrank} that reduces the embedding move distance is expected to
be resistant to our OA to some extent.
To validate this, we conduct white-box attack on a defensed model
as shown in Tab.~\ref{tab:white-def}, 
as well as black-box attack (with SPSA) on a defensed model
as shown in Tab.~\ref{tab:black-def}.

\begin{table}[h]
\resizebox{1.0\columnwidth}{!}{
\setlength{\tabcolsep}{0.5em}
\begin{tabular}{c|cc|cc|cc|cc|cc|cc}
\toprule
\multirow{3}{*}{$\varepsilon$} & \multicolumn{6}{c|}{\cellcolor{lime!15}Fashion-MNIST ($N=\infty$)} & \multicolumn{6}{c}{\cellcolor{cyan!15}Stanford Online Product ($N=\infty$)}\tabularnewline
\cline{2-13} \cline{3-13} \cline{4-13} \cline{5-13} \cline{6-13} \cline{7-13} \cline{8-13} \cline{9-13} \cline{10-13} \cline{11-13} \cline{12-13} \cline{13-13}
 & \multicolumn{2}{c|}{$k=5$} & \multicolumn{2}{c|}{$k=10$} & \multicolumn{2}{c|}{$k=25$} & \multicolumn{2}{c|}{$k=5$} & \multicolumn{2}{c|}{$k=10$} & \multicolumn{2}{c}{$k=25$}\tabularnewline
\cline{2-13} \cline{3-13} \cline{4-13} \cline{5-13} \cline{6-13} \cline{7-13} \cline{8-13} \cline{9-13} \cline{10-13} \cline{11-13} \cline{12-13} \cline{13-13} 
 & $\tau_\mathsfit{S}$ & mR & $\tau_\mathsfit{S}$ & mR & $\tau_\mathsfit{S}$ & mR & $\tau_\mathsfit{S}$ & mR & $\tau_\mathsfit{S}$ & mR & $\tau_\mathsfit{S}$ & mR\tabularnewline
\midrule
2/255 & 0.098 & 2.0 & 0.069 & 4.6 & 0.033 & 12.1 & 0.216 & 2.1 & 0.190 & 4.7 & 0.093 & 12.4\tabularnewline
\hline 
4/255 & 0.176 & 2.1 & 0.122 & 4.7 & 0.059 & 12.3 & 0.306 & 2.4 & 0.294 & 5.1 & 0.148 & 13.0\tabularnewline
\hline 
8/255 & 0.271 & 2.4 & 0.207 & 4.9 & 0.096 & 12.6 & 0.381 & 2.6 & 0.380 & 5.9 & 0.206 & 13.8\tabularnewline
\hline 
16/255 & 0.384 & 3.1 & 0.304 & 5.8 & 0.137 & 13.4 & 0.526 & 3.2 & 0.433 & 6.8 & 0.249 & 14.7\tabularnewline
\bottomrule
\end{tabular}
}
\caption{White-Box attack against a defensed model~\cite{advrank}.}
\label{tab:white-def}
\end{table}

\begin{table}[h]
\resizebox{1.0\columnwidth}{!}{
\setlength{\tabcolsep}{0.3em}
\begin{tabular}{c|ccc|ccc|ccc|ccc}
\toprule
\multirow{3}{*}{$\varepsilon$} & \multicolumn{6}{c|}{\cellcolor{lime!15}Fashion-MNIST } & \multicolumn{6}{c}{\cellcolor{cyan!15}Stanford Online Product}\tabularnewline
\cline{2-13} \cline{3-13} \cline{4-13} \cline{5-13} \cline{6-13} \cline{7-13} \cline{8-13} \cline{9-13} \cline{10-13} \cline{11-13} \cline{12-13} \cline{13-13} 
 & \multicolumn{3}{c|}{$N=\infty$} & \multicolumn{3}{c|}{$N=50$} & \multicolumn{3}{c|}{$N=\infty$} & \multicolumn{3}{c}{$N=50$}\tabularnewline
\cline{2-13} \cline{3-13} \cline{4-13} \cline{5-13} \cline{6-13} \cline{7-13} \cline{8-13} \cline{9-13} \cline{10-13} \cline{11-13} \cline{12-13} \cline{13-13} 
	& $k=5$ & $k=10$ & $k=25$ & $k{=}5$ & $k{=}10$ & \multicolumn{1}{c|}{$k{=}25$} & $k=5$ & $k=10$ & $k=25$ & $k{=}5$ & $k{=}10$ & $k{=}25$\tabularnewline
\midrule
2/255 & 0.114, 2.0 & 0.099, 4.6 & 0.052, 12.1 & 0.113 & 0.099 & 0.050 & 0.057, 2.0 & 0.058, 4.5 & 0.041, 12.0 & 0.058 & 0.061 & 0.042\tabularnewline
\hline 
4/255 & 0.154, 2.1 & 0.156, 4.7 & 0.090, 12.2 & 0.156 & 0.158 & 0.089 & 0.075, 2.0 & 0.109, 4.6 & 0.086, 12.1 & 0.075 & 0.102 & 0.082\tabularnewline
\hline 
8/255 & 0.172, 2.2 & 0.217, 4.9 & 0.138, 12.8 & 0.178 & 0.216 & 0.131 & 0.082, 2.0 & 0.133, 4.7 & 0.128, 12.4 & 0.073 & 0.138 & 0.131\tabularnewline
\hline 
16/255 & 0.177, 2.3 & 0.259, 5.4 & 0.167, 13.8 & 0.179 & 0.259 & 0.165 & 0.068, 2.1 & 0.162, 4.9 & 0.158, 13.2 & 0.069 & 0.159 & 0.162\tabularnewline
\bottomrule
\end{tabular}
}
\caption{Black-Box attack against a defensed model~\cite{advrank}.}
\label{tab:black-def}
\end{table}

According to the tables, the defense is moderately effective against OA.
Further analysis is left for future work.

\section{Difference from Absolute Rank Attack}

The equations in Sec.~\ref{sec:white} are similar to those
in \cite{advrank} to make the method self-contained.
Although visually similar to QA+~\cite{advrank},
$L_\text{ReO}$ in Eq.~\ref{eq:reo} has a
different goal:\\
\includegraphics[width=1.0\columnwidth]{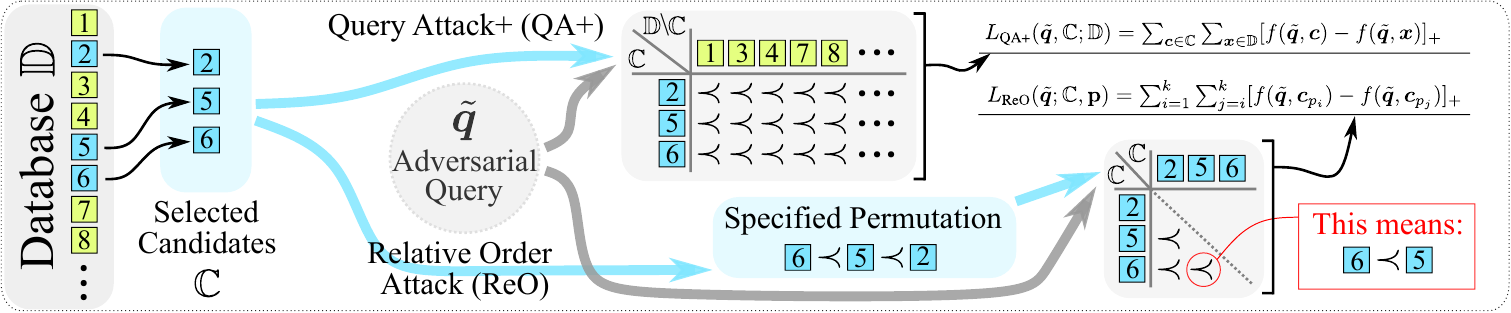}
Besides, we also contribute a surrogate objective (Short-range Ranking Correlation)
in Sec.~3.2 for black-box OA.

\section{Application \& Influence of OA}


On popular online shopping platforms, sales of a product closely correlates
to the click-through rate (CTR).
Furthermore, the ranking position has significant impacts on its CTR,
per sponsored search advertising
literature~\cite{chen2012position,pin2011stochastic}.
Specifically, ``the
observed CTR is geometrically decreasing as the position lowers down,
exhibiting a good fit to the gamma signature.''~\cite{chen2012position}.
In practice, $\text{Gamma}(1.2,0.01)$ distribution is used to model this
position effect, whose value is highly sensitive to even subtle changes in
ranking positions.
Take the Taobao.com\footnote{An online shopping site owned by Alibaba (NYSE:BABA).}
(mobile version) as an example, only 4 slots are typically shown in each page
of retrieval with the first being paid promotion.
For advertisers, the cost difference between the 1st page Ads and 2nd page Ads 
is dramatic,
which indicates the monetary
value of this attack (\eg, moving a good's position from 4th to 3th leads to
its appearance on the 1st page of retrieval results).

	To help the reader better understand the motivation and potential
	influence of OA, we elaborate on a concrete example about how
	it may be used in practice.

	Recall a claim in the Introduction of the manuscript:
	``{\it
	Such vulnerability in a commercial platform may
	be exploited in a malfeasant business competition among
	the top-ranked products, \eg, via promoting fraudulent web
	pages containing adversarial example product images.}''
	Assume product A is the best seller of its kind.
	Product B and C are A's alternatives and business competitors, but are less prevalent.
	%
	%
	When a client searches with product image of A, product A, B, and C will be ranked near the
	topmost part of the list and presented to the client, while A is ranked ahead of B and C.
	Then, an attacker may want to to increase B's and C's sales leveraging A's popularity
	and advertising effects.
	Product C may have a higher priority than B because
	the producer of C invested more money to the attacker.
	
	An attacker may first
	setup a third-party product promotion website displaying images
	of product A, but these images are actually adversarially perturbed
	with \emph{Order Attack} to make B and C ranked ahead of A, and C ahead of B
	without introducing obviously wrong
	retrieval results.
	Namely, this website is pretending
	to promote product A, but is actually promoting product C and B as their
	\emph{relative order} has been changed into C $\prec$ B $\prec$ A.
	When a user clicks an adversarially perturbed product image of A, the
	image is used as a query, and
	the user will first see product C, then B, finally A.
	Such subtle changes in \emph{relative order} can be sufficient to
	impact the relative sales of product A, B and C.
	%
	%
	Thus, this is an example of ``malfeasant business competition''
	which cannot be achieved by pure \emph{absolute rank} attacks.
	Ways to guide users to such websites, \eg, phishing or
	hijacking user requests, are beyond the scope of discussion.


	\newpage
	\begin{figure*}[t]
		\includegraphics[width=1.0\textwidth]{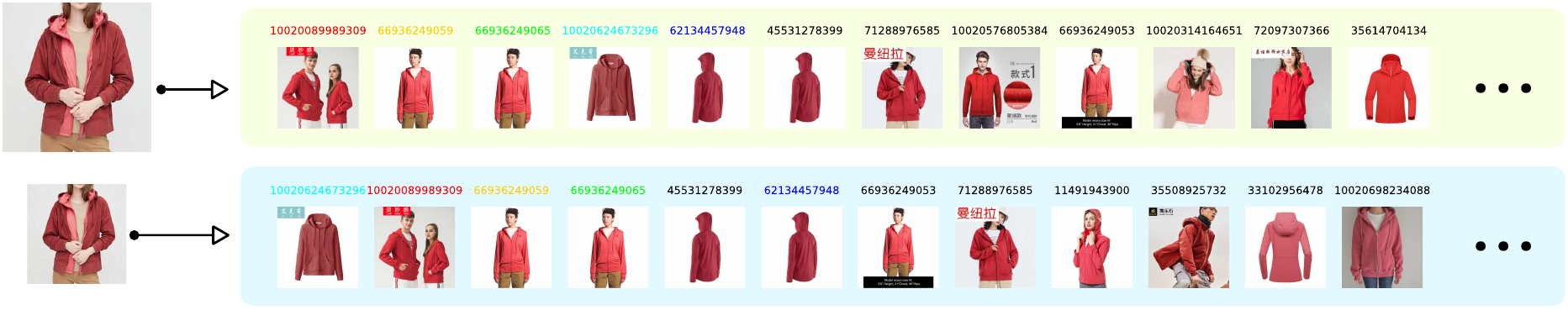}
		\caption{Showcase \#2: ``Red wind coat'' query image with $\varepsilon=2/255$ perturbation.}
		\label{fig:advorder2}
	\end{figure*}

	\begin{figure*}[t]
		\includegraphics[width=1.0\textwidth]{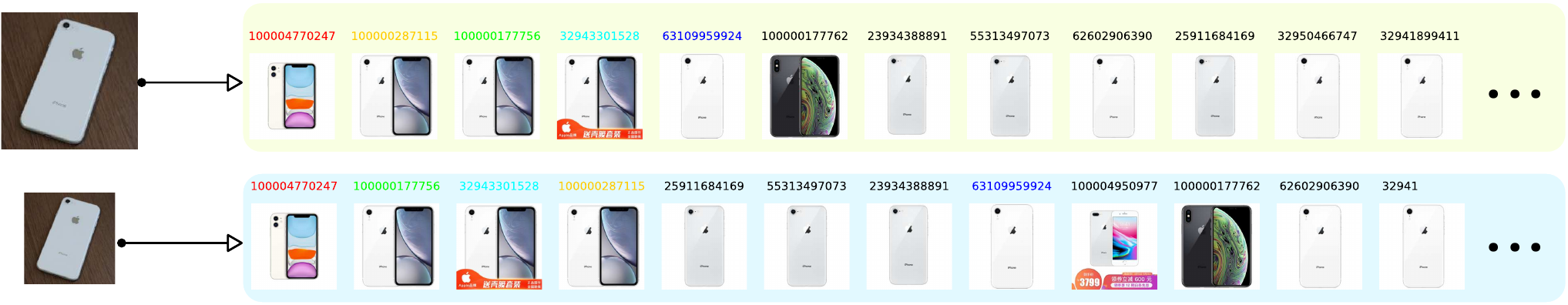}
		\caption{Showcase \#3: ``iPhone'' query image with $\varepsilon=1/255$ perturbation.}
		\label{fig:advorder3}
	\end{figure*}

	\begin{figure*}[t]
		\includegraphics[width=1.0\textwidth]{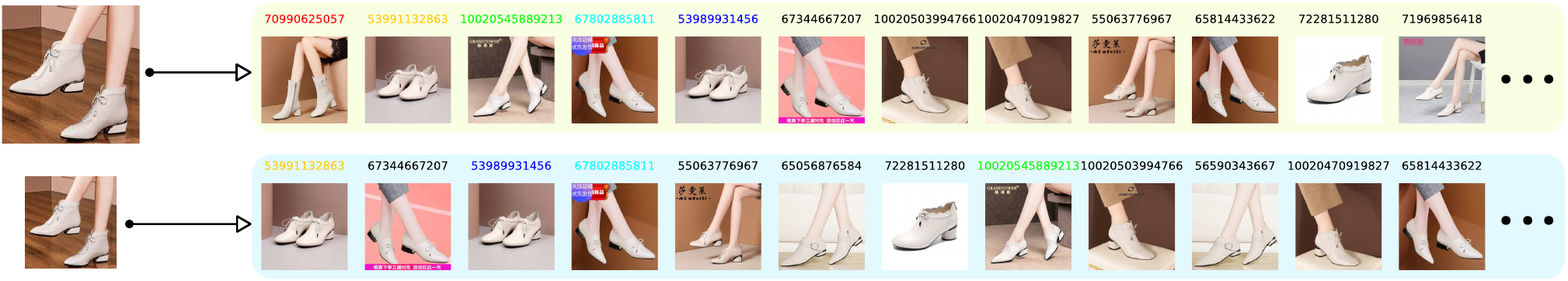}
		\includegraphics[width=1.0\textwidth]{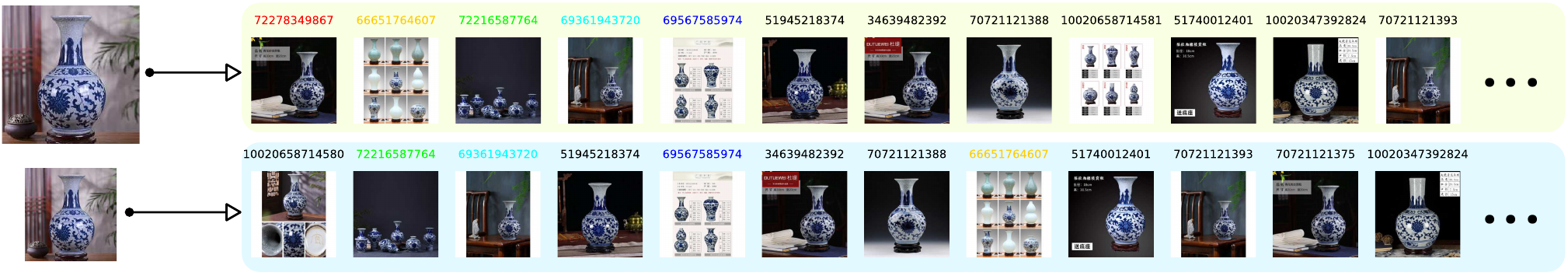}
		\caption{Two cases (``Shoe'' and ``Vase'') where the top-$1$ ranked candidate disappear from the top-ranked results.}
		\label{fig:miss1}
	\end{figure*}

	\begin{figure*}[t]
		\includegraphics[width=1.0\textwidth]{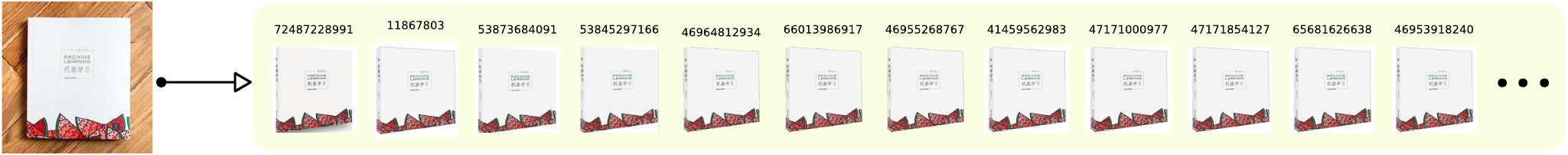}
		\includegraphics[width=1.0\textwidth]{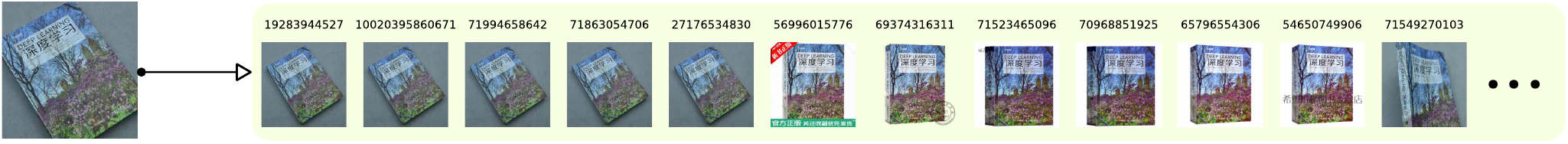}
		\includegraphics[width=1.0\textwidth]{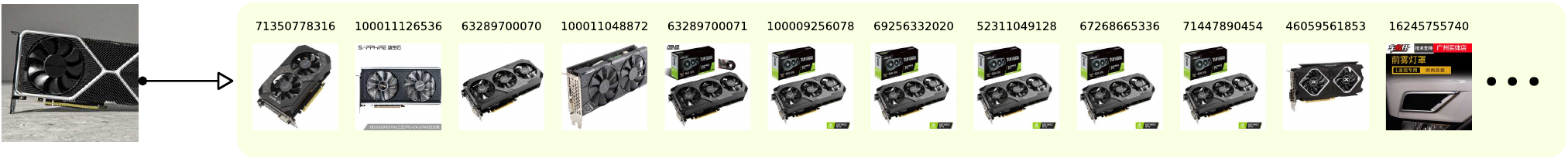}
		\caption{Three long-tail query cases (Two ``Textbooks'' and ``Graphics Card'') where OA will not be effective as expected.}
		\label{fig:longtail}
	\end{figure*}
	\clearpage

	\newpage
	\begin{figure*}[t]
		\centering
		\begin{minipage}{0.48\textwidth}
			\includegraphics[width=1.0\textwidth]{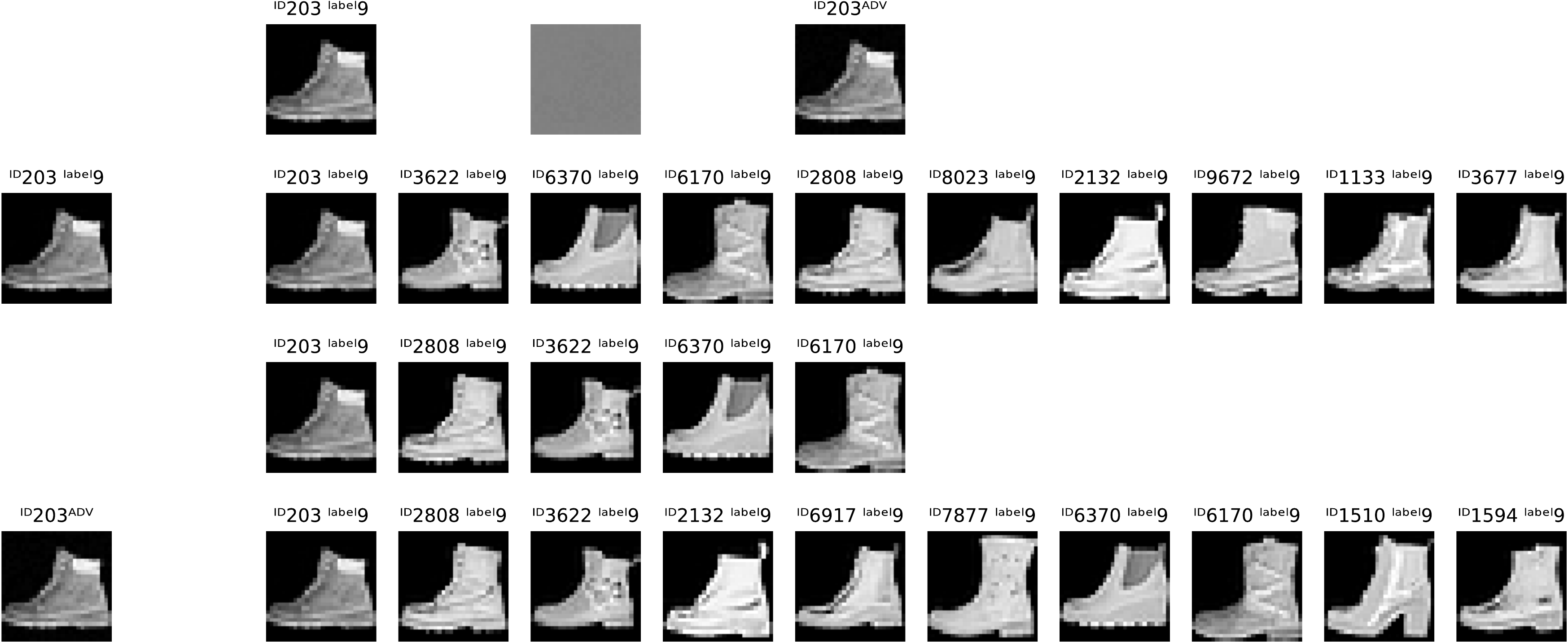}
			\caption{Fashion-MNIST Showcase \#1. $k=5$, $\taun=1.0$}
			\label{fig:fb1}
		\end{minipage}
		\hfill \vline \hfill
		\begin{minipage}{0.48\textwidth}
			\includegraphics[width=1.0\textwidth]{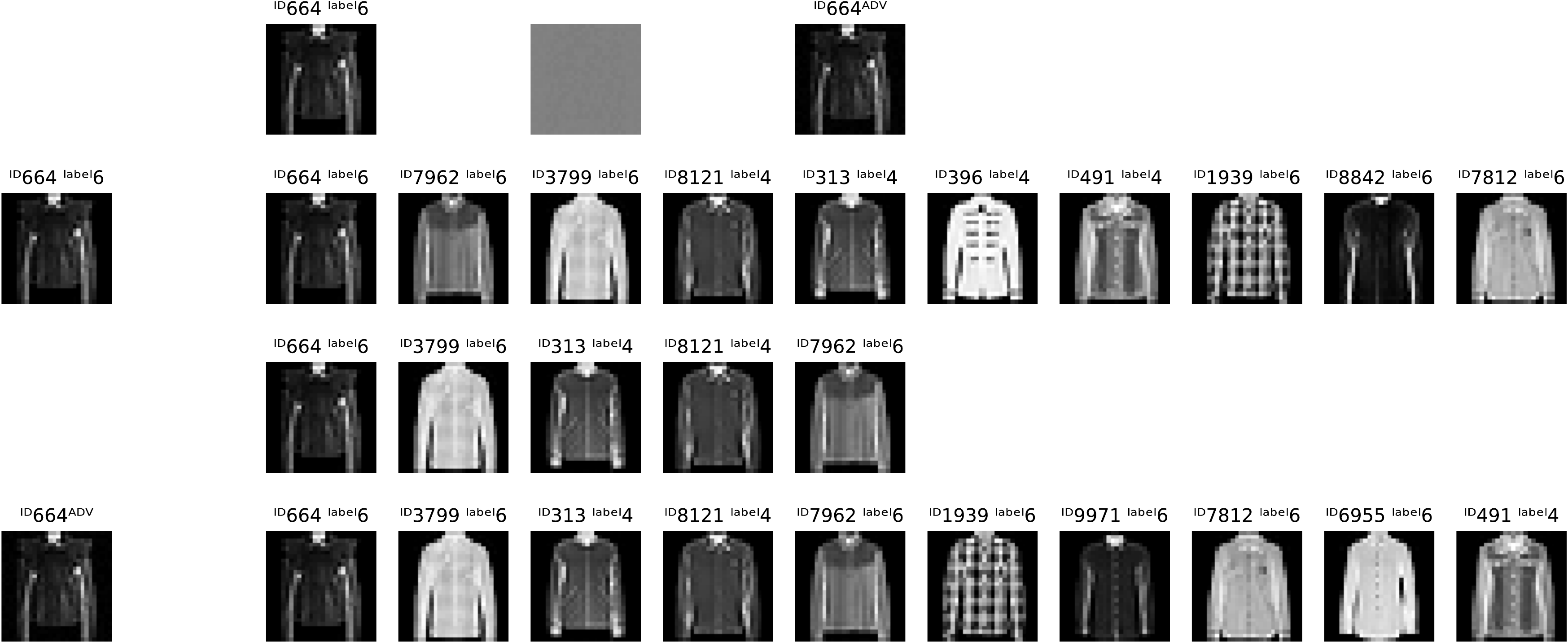}
			\caption{Fashion-MNIST Showcase \#2. $k=5$, $\taun=1.0$}
			\label{fig:fb2}
		\end{minipage}
		\newline
		\begin{minipage}{1.0\textwidth}
			\includegraphics[width=1.0\textwidth]{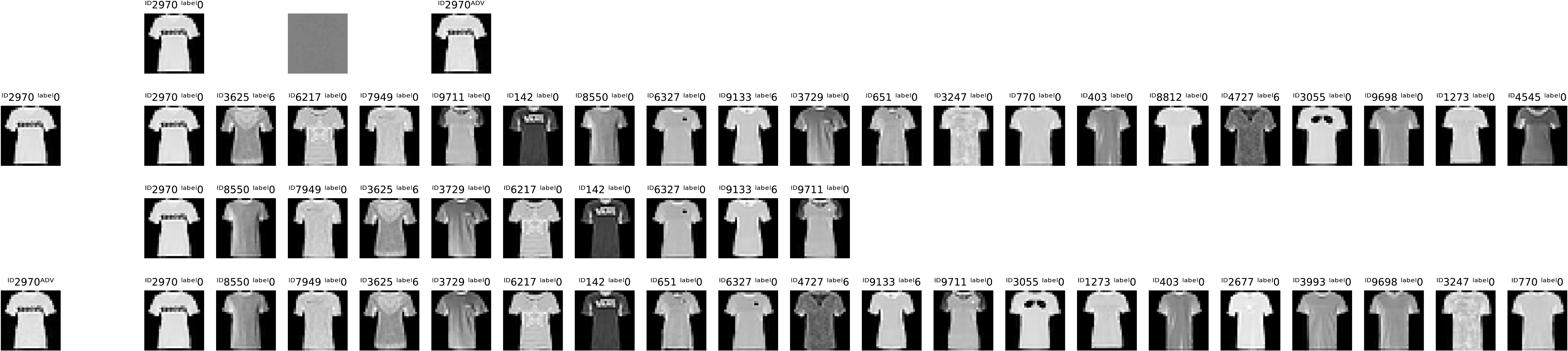}
			\caption{Fashion-MNIST Showcase \#3. $k=10$, $\taun=1.0$}
			\label{fig:fb3}
		\end{minipage}
	\end{figure*}

	\begin{figure*}[t]
		\centering
		\begin{minipage}{0.48\textwidth}
			\includegraphics[width=1.0\textwidth]{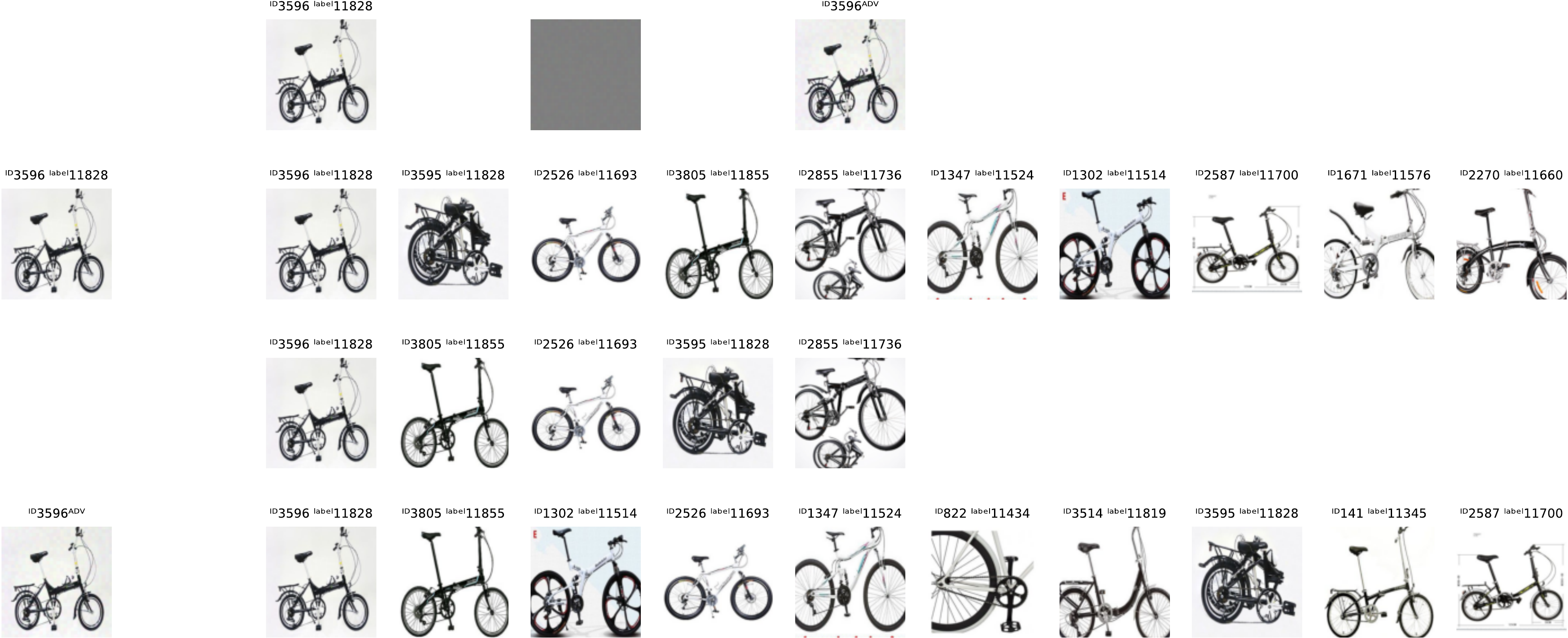}
			\caption{SOP Showcase \#1. $k=5$, $\taun=1.0$}
			\label{fig:sb1}
		\end{minipage}
		\hfill \vline \hfill
		\begin{minipage}{0.48\textwidth}
			\includegraphics[width=1.0\textwidth]{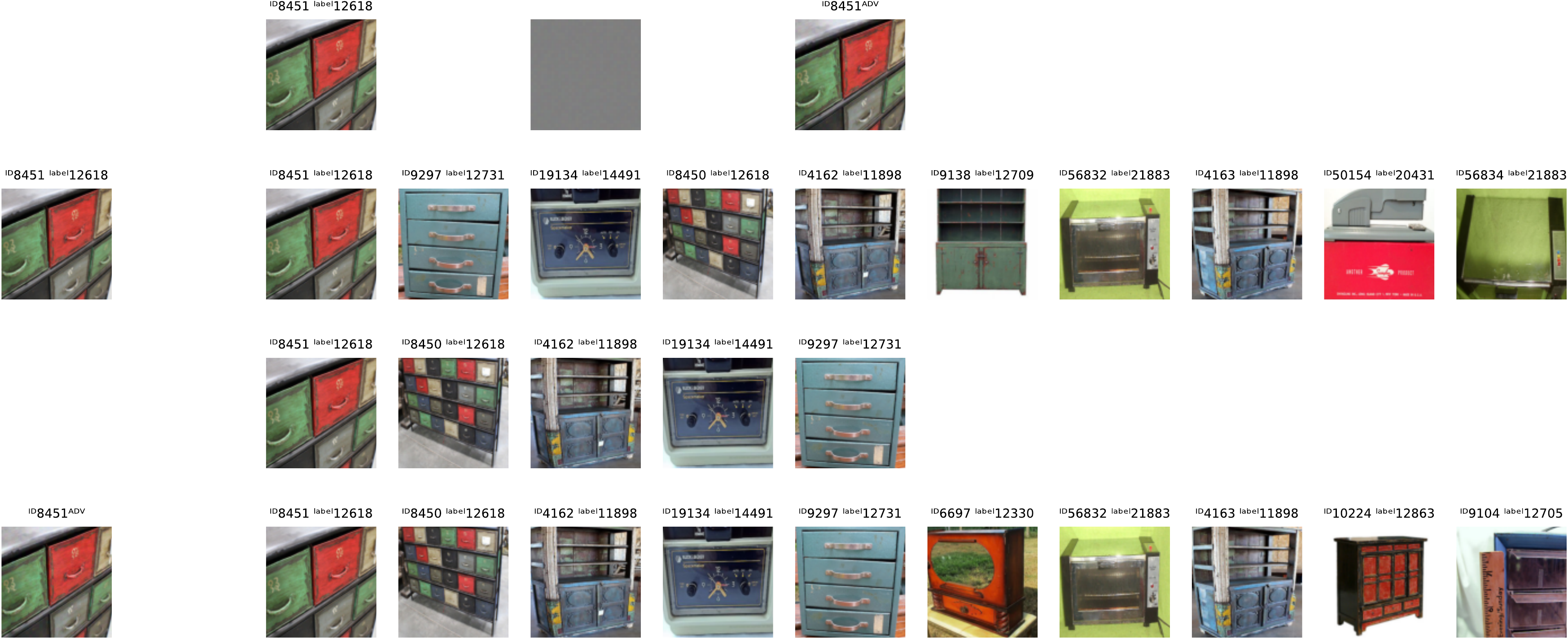}
			\caption{SOP Showcase \#2. $k=5$, $\taun=1.0$}
			\label{fig:sb2}
		\end{minipage}
		\newline
		\begin{minipage}{1.0\textwidth}
			\includegraphics[width=1.0\textwidth]{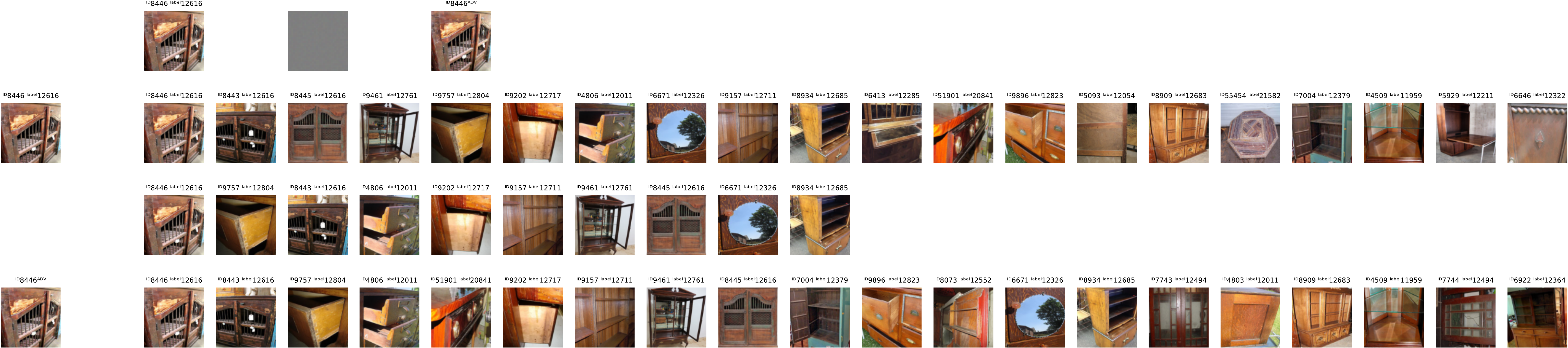}
			\caption{SOP Showcase \#3. $k=10$, $\taun=0.96$}
			\label{fig:sb3}
		\end{minipage}
	\end{figure*}
	\clearpage


\newpage
\section{Black-Box Optimizer Details}

In this section, we present the black-box optimization algorithm details for
(1) Random Search (Rand);
(2) Beta-Attack (Beta);
(3) Particle Swarm Optimization (PSO)~\cite{stdPSO};
(4) Natural Evolution Strategy (NES)~\cite{nes-atk,nes-algo};
and (5) Simultaneous Perturbation Stochastic Approximation (SPSA)~\cite{spsa-atk,spsa-algo}.
Experimental results of parameter search for every optimization algorithm
are also provided.

\subsection{Random Search (Rand)}

As a baseline algorithm for black-box optimization, 
Random Search assumes each element in
the adversarial perturbation to be \emph{i.i.d}, and samples each element from
the uniform distribution within $\Omega_\vq$, namely
$\vr=[r_1,r_2,\ldots,r_D]$ where $r_i \sim \mathcal{U}(-\varepsilon,+\varepsilon)$ ($i=1,2,\ldots,D$)
in each iteration.
The output of the algorithm is the best historical result, as summarized in
Algo.\ref{algo:rand}.
This algorithm is free of hyper-parameters.

This algorithm will never be stuck at a local-maxima, which means it has a great
ability to search for solutions from the global scope.
But its drawback
is meanwhile clear, as each trial of this algorithm is independent to each other.
In our implementation, we conduct OA on a batch of random perturbations
to accelerate the experiments, with the batch size set as $H=50$.

\begin{algorithm}[h]
	\SetAlgoLined
	\KwIn{Query Image $\vq$, Query Budget $Q$, Selected Candidates $\sC$, Permutation vector $\rvp$}
	\KwOut{Adversarial Query $\tilde{\vq}$}
	Initialize $\tilde{\vq} \gets \vq$, and the best score $s^\ast \gets \taun(\tilde{\vq})$\;
	\For{$i \gets 1,2,\ldots,Q$}{
		Sample $\vr \sim \mathcal{U}^D(-\varepsilon,+\varepsilon)$\;
		\If{$\taun\Big(\text{\upshape Clip}_{\Omega_\vq}(\vq + \vr)\Big) > s^\ast$}{
			$\tilde{\vq} \gets \text{\upshape Clip}_{\Omega_\vq}(\vq + \vr)$ \;
			$s^\ast \gets \taun\Big(\text{\upshape Clip}_{\Omega_\vq}(\vq + \vr)\Big)$ \;
		}
	}
	\Return{$\tilde{\vq}$}
	\caption{Rand: Na\"{i}ve Random Search.}
	\label{algo:rand}
\end{algorithm}

	\subsection{Beta-Attack (Beta)}

Beta-Attack is modified from $\mathcal{N}$-Attack~\citep{nattack}.
Although similar, a notable
difference between them is that the Gaussian distributions in $\mathcal{N}$-Attack
are replaced with Beta distributions.
We choose Beta distribution because
the shape of its probability density function is much more flexible than
that of the Gaussian distribution, which may be beneficial for modeling
the adversarial perturbations.
Besides, according to our observation,
$\mathcal{N}$-Attack is too prone to be stuck at a local maxima for OA,
leading to a considerably low $\taun$.

The key idea of Beta-Attack is to find the parameters for a parametric
distribution $\pi(\vz|\bm{\theta})$ from which the adversarial perturbations drawn
from it is likely adversarially effective.
For Beta-Attack, $\pi(\vz|\bm{\theta})$ is a combination of $D$ independent
Beta distributions \footnote{The multivariate Beta distribution,
\emph{a.k.a} Dirichlet distribution $\vz\sim\text{Dir}(\va)$
is not used here because its $\sum z_i=1$ restriction further shrinks
the search space hence may lower the upper-bound of $\taun$.},
with parameter $\bm{\theta}=[\va;\vb]$ where
$\va=[a_1,a_2,\ldots,a_D]$, $\vb=[b_1,b_2,\ldots,b_D]$, and $z_i\sim \text{Beta}(a_i,b_i) \in [0,1], ~(i=1,2,\ldots,D)$.
Let $\mathcal{T}(\vz)=\taun\big(\text{Clip}_{\Omega_\vq}(\vq + \varepsilon(2\vz-1))\big)$,
where $\varepsilon(2\vz-1)=\vr$ is the adversarial perturbation.
We hope to maximize the mathematical expectation of $\mathcal{T}(\vz)$ over distribution $\pi(\vz|\bm{\theta})$:
\begin{equation}
	\max_\mathbf{\theta} \E_{\pi(\vz|\bm{\theta})} \Big[ \mathcal{T}(\vz) \Big] :=
	\int \mathcal{T}(\vz) \cdot \pi(\vz|\bm{\theta}) d\vz
\end{equation}
The gradient of the expectation with respect to $\bm\theta$ is
\begin{align}
	& \nabla_{\bm{\theta}} \E_{\pi(\vz|\bm{\theta})} \Big[ \mathcal{T}(\vz) \Big] \\
	=&
	\nabla_{\bm{\theta}} \int \mathcal{T}(\vz) \cdot \pi(\vz|\bm{\theta}) d\vz \label{eq:6}\\
	=& \int \mathcal{T}(\vz) \cdot \frac{\pi(\vz|\bm{\theta})}{\pi(\vz|\bm{\theta})}
	\nabla_{\bm{\theta}} \pi(\vz|\bm{\theta}) d\vz\\
	=& \int \pi(\vz|\bm{\theta}) \cdot \mathcal{T}(\vz) \cdot \nabla_{\bm{\theta}} \log \big[\pi(\vz|\bm{\theta}) \big]d\vz\\
	=& \E_{\pi(\vz|\bm{\theta})} \Big[ \mathcal{T}(\vz) \cdot \nabla_{\bm{\theta}} \log \big[\pi(\vz|\bm{\theta}) \big] \Big] \label{eq:9}
\end{align}
where
\begin{small}
\begin{align}
	\nabla_{\bm{\theta}} \log \big[\pi(\vz|\bm{\theta}) \big] &=
	\Big[ \nabla_\va \log \big[\pi(\vz|\bm{\theta}) \big];~
	\nabla_\vb \log \big[\pi(\vz|\bm{\theta}) \big]  \Big], \\
	\nabla_\va \log \big[\pi(\vz|\bm{\theta}) \big] &=
		\psi^{(0)}(\va + \vb) - \psi^{(0)}(\va) + \log(\vz),\\
	\nabla_\vb \log \big[\pi(\vz|\bm{\theta}) \big] &=
		\psi^{(0)}(\va + \vb) - \psi^{(0)}(\vb) + \log(1-\vz),\label{eq:12}
\end{align}
\end{small}
and $\psi^{(n)}(z)$ is the $n$-th derivative of the digamma function.
The Eq.~\ref{eq:6} to Eq.~\ref{eq:12} means that the gradient of the expectation
of $\mathcal{T}(\vz)$ with respect to $\theta$ can be estimated by approximating
the expectation in Eq.~\ref{eq:9} with its mean value using
a batch of random vectors, \ie,
\begin{align}
	& \E_{\pi(\vz|\bm{\theta})} \Big[ \mathcal{T}(\vz) \cdot \nabla_{\bm{\theta}} \log \big[\pi(\vz|\bm{\theta}) \big] \Big] \\
	\approx&
	\frac{1}{H} \sum_{i=1}^{H} \Big[ \mathcal{T}(\vz_i) \cdot \nabla_{\bm{\theta}} \log \big[\pi(\vz_i|\bm{\theta}) \big] \Big]
\end{align}
where $H$ denotes the batch size, and $\vz_i$ is drawn from $\pi(\vz|\bm\theta)$.
Thus, the parameters $\bm\theta$ of the Beta distributions can be updated 
with Stochastic Gradient Ascent, \ie,
\begin{equation}
{\bm\theta}_{t+1} \leftarrow \bm{\theta}_t + \eta \nabla_{\bm\theta} \E_{\pi(\vz|\mathbf{\bm\theta})} \Big[ \mathcal{T}(\vz) \Big],
\end{equation}
where $\eta$ is a constant learning rate for the parameters $\bm\theta$.
With a set of trained parameters $\bm\theta$, we expect a higher $\taun$ performance
from a random perturbation $\vz$ drawn from $\pi(\vz|\bm\theta)$.

In our experiments, we initialize $\va{=}1$, and $\vb{=}1$.
This is due to a important property of Beta distribution that
it degenerates into Uniform distribution when $a{=}1$ and $b{=}1$.
Namely, our Beta-Attack is initialized as the ``Rand Search'' method, but it is able
to update its parameters according to the historical $\taun$ results,
changing the shape of its probability density function in order to improve the expectation
of the $\taun$ of the next adversarial perturbation drawn from it.
The batch size $H$ is set to $50$ for all the experiments.

\ifsupp
As discussed in Sec.~4.2, and shown in
Tab.~3 and Tab.~4, the Beta-Attack
\else
As discussed in Sec.~\ref{sec:expblack}, and shown in
Tab.~\ref{tab:black-fashion} and Tab.~\ref{tab:black-sop}, the Beta-Attack
\fi
obviously outperforms the ``Rand Search'', and is comparable to PSO, but is still
surpassed by NES and SPSA.
Such simple parametric distributions are far not enough
for modeling the adversarial perturbations for the challenging OA problem.
Due to its performance not outperforming PSO, NES and SPSA,
the new Beta-Attack is not regarded as a
contribution of our paper, but its comparison with the other methods are
still very instructive.

\subsubsection{Parameter Search for Beta-Attack}


\begin{table}[h]
\centering
\resizebox{1.0\columnwidth}{!}{
\setlength{\tabcolsep}{0.5em}%
\begin{tabular}{c|ccccc}
	\toprule
\rowcolor{lime!10}Learning Rate $\eta$ & 0.0 & 1.5 & {*}3.0 & 4.5 & 6.0\tabularnewline
\midrule
SRC $\taun$ & 0.290, 2.2 & 0.332, 2.4 & 0.360, 2.6 & 0.341, 2.5 & 0.330, 2.5\tabularnewline
\bottomrule
\end{tabular}
}
\caption{Parameter search of learning rate $\eta$ for Beta-Attack. Fashion-MNIST dataset, $N=\infty$, $k=5$, $\varepsilon=\frac{4}{255}$.}
\label{tab:parabeta}
\end{table}


We conduct parameter search of $\eta$ on the Fashion-MNIST, as shown in
Tab.~\ref{tab:parabeta}. From the table, we find that the $\taun$ performance
peaks at $\eta=3.0$, so we set this value as the default learning rate
for the experiments on Fashion-MNIST. Apart from that, we empirically
set $\eta=0.5$ for the experiments on SOP following a similar parameter search.

	\subsection{Particle Swarm Optimization (PSO)}

Particle Swarm Optimization~\citep{PSO,stdPSO} is a classical meta-heuristic optimization
method. Let constant $H$ denote the population (swarm) size, we randomly initialize the
$H$ particles (vectors) as $\sY=\{\vy_1, \vy_2, \ldots, \vy_H\}$. The positions
of these particles are iteratively updated according to the following velocity formula:
\begin{equation}
	\vv_i  \leftarrow \omega \vv_i + \text{rand()} \cdot \phi_p (\vp_i - \vy_i)
	+ \text{rand()} \cdot \phi_g (\vg - \vy_i)\\
\end{equation}
\begin{equation}
	\vy_i  \leftarrow \vy_i + \vv_i
\end{equation}
where $\vy_i\in \sY$, $\text{rand()}$ generates a random number within the interval $[0,1]$,
$\omega$ denotes the inertia (momentum), $\vp_i$ is the historical best position
of particle $i$, $\vg$ is the global historical best position among all particles,
$\phi_p$ and $\phi_g$ are two constant parameters.
As a meta-heuristic method, PSO does not guarantee that a satisfactory solution
will eventually be discovered.

In the implementation, we directly represent the adversarial query $\tilde{\vq}$
withe the particles $\sY$. And we also additionally clip the particles
at the end of each PSO iteration, \ie,
\begin{equation}
\vy_i\leftarrow \min\Big\{ \vq+\varepsilon, \max\big\{ \vq-\varepsilon, \vy_i \big\} \Big\},
\end{equation}
which is the only difference of our implementation
compared to the standard PSO~\citep{stdPSO}.
In all our experiments, the swarm size is empirically set as $H=40$.

\subsubsection{Parameter Search for PSO}

\begin{table}[h]
\centering
\resizebox{1.0\columnwidth}{!}{
\setlength{\tabcolsep}{0.5em}%
\begin{tabular}{c|ccccc}
	\toprule
\rowcolor{lime!10}Inertia $\omega$ & 0.8 & 1.0 & {*}1.1 & 1.2 & 1.4\tabularnewline
\midrule
SRC $\taun$ & 0.321, 2.3 & 0.363, 2.3 & 0.381, 2.3 & 0.369, 2.4 & 0.349, 2.4\tabularnewline
\bottomrule
\end{tabular}
	}
	\caption{Inertia $\omega$ parameter search for PSO. Fashion-MNIST dataset, $N=\infty$, $k=5$, $\varepsilon=\frac{4}{255}$.}
	\label{tab:parapso2}
\end{table}

\begin{table}[h]
	\centering
	\resizebox{1.0\columnwidth}{!}{
	\setlength{\tabcolsep}{0.5em}%
	\begin{tabular}{c|ccccc}
	\toprule
\rowcolor{lime!10}$\phi_p$ & 0.37 & 0.47 & {*}0.57 & 0.67 & 0.77\tabularnewline
	\midrule
SRC $\taun$ & 0.353, 2.3 & 0.375, 2.3 & 0.381, 2.3 & 0.376, 2.3 & 0.364, 2.3\tabularnewline
	\bottomrule
\end{tabular}
}
\caption{$\phi_p$ parameter search for PSO. Fashion-MNIST dataset, $N=\infty$, $k=5$, $\varepsilon=\frac{4}{255}$.}
	\label{tab:parapso3}
\end{table}

\begin{table}[h]
\centering
	\resizebox{1.0\columnwidth}{!}{
	\setlength{\tabcolsep}{0.5em}%
	\begin{tabular}{c|ccccc}
		\toprule
\rowcolor{lime!10}$\phi_g$ & 0.24 & 0.34 & {*}0.44 & 0.54 & 0.64\tabularnewline
\midrule
SRC $\taun$ & 0.353, 2.3 & 0.372, 2.3 & 0.381, 2.3 & 0.380, 2.3 & 0.359, 2.3\tabularnewline
\bottomrule
\end{tabular}
}
\caption{$\phi_g$ parameter search for PSO. Fashion-MNIST dataset, $N=\infty$, $k=5$, $\varepsilon=\frac{4}{255}$.}
\label{tab:parapso}
\end{table}


We conduct parameter search of $\omega$, $\phi_p$ and $\phi_g$ for PSO on the
Fashion-MNIST dataset, as shown in Tab.~\ref{tab:parapso2}, Tab.~\ref{tab:parapso3} and Tab.~\ref{tab:parapso}.
\begin{itemize}
	\item Inertia $\omega$: This parameter affects the convergence of the algorithm.
		A large $\omega$ endows PSO with better global searching ability, while a
		small $\omega$ allows PSO to search better in local areas. From the
		table, we find the PSO performance peaks at $\omega=1.1$.
		The performance curve of $\taun$ with respect to different $\omega$ settings suggests that
		the global searching ability is relatively important for solving
		the black-box Order Attack problem.
	\item Constants $\phi_p$ and $\phi_g$: These parameter control the weights
		of a particle's historical best position (``the particle's own knowledge'')
		and the swarm's historical global best position (``the knowledge shared
		among the swarm'') in the velocity formula. With a relatively small $\phi_p$,
		and a relatively large $\phi_g$, the swarm
		will converge faster towards the global best position, taking a higher
		risk of being stuck at a local optimum.
		From the tables of parameter search, we note that
		both constants should be kept relatively small, which means it is
		not preferred to converge too fast towards either the particles' individual
		best positions or the global best position, for sake of better global
		searching ability of the algorithm.
\end{itemize}
We conclude from the parameter search that the global searching ability
is important for solving the black-box OA problem, as reflected by the parameter search.
Hence, we empirically set $\omega=1.1$, $\phi_p=0.57$, and $\phi_g=0.44$
for PSO in all experiments.

	\subsection{Natural Evolutionary Strategy (NES)}

NES~\citep{nes-atk} is based on \citet{nes-algo}, which aims to find the adversarial
perturbation through projected gradient method \citep{madry} with estimated gradients.
Specifically, let the adversarial query $\tilde{\vq}$ be the mean of a multivariate
Gaussian distribution $\mathcal{N}(\vz|\tilde{\vq},\bm{\Sigma})$ where the covariance
matrix $\bm\Sigma$ is a hyper-parameter matrix, and
$\mathcal{T}(\vz) = \taun\big(\text{Clip}_{\Omega_\vq}(\tilde{\vq} + \vz) \big)$,
and $\tilde{\vq}$ is initialized as $\vq$.
NES first estimates the gradient of the expectation with a batch of random
points sampled near the adversarial query:
\begin{align}
	& \nabla_{\tilde{\vq}} \E_{\mathcal{N}(\vz|\tilde{\vq},\bm{\Sigma)}} [\mathcal{T}(\vz)] \\
	=~& \E_{\mathcal{N}(\vz|\tilde{\vq},\bm{\Sigma})} \Big\{\mathcal{T}(\vz) \nabla_{\tilde{\vq}} [
		\log \mathcal{N}(\vz|\tilde{\vq},\bm{\Sigma})] \Big\}\\
	\approx ~&
	\frac{1}{H} \sum_{i=1}^H \Big\{ \mathcal{T}(\vz_i) \nabla_{\tilde{\vq}} [
		\log \mathcal{N}(\vz_i|\tilde{\vq},\bm{\Sigma})] \Big\}.
\end{align}
The batch of $\vz_i$ are sampled from $\mathcal{N}(\tilde{\vq},\bm{\Sigma})$.
Then NES updates the adversarial example $\tilde{\vq}$ (\ie, the mean of the 
multivariate Gaussian) with Projected Gradient
Ascent~\citep{madry}:
\begin{equation}
	\tilde{\vq}_{t+1} \leftarrow
	\text{Clip}_{\Omega_\vq}\Big\{\tilde{\vq_t} + \eta \cdot \text{sign}\big(
	\nabla_{\tilde{\vq}} \E_{\mathcal{N}(\vz|\tilde{\vq},\bm{\Sigma})} [\mathcal{T}(\vz)]
	\big) \Big\},
\end{equation}
where $\eta$ is a constant learning rate.
Following \citet{nes-atk}, we set the covariance matrix as a scaled 
identity matrix, \ie, $\Sigma=\sigma I$. The batch size is set to $H=50$
for all experiments. In other words, each step of Projected Gradient Ascent
will be based on the gradient estimated using a batch of $H=50$ random
vectors ($\vz_1, \vz_2, \ldots, \vz_H$) drawn from $\mathcal{N}(\tilde{\vq},\bm{\Sigma})$.

		\subsubsection{Parameter Search for NES}

\begin{table}[h]
\centering
\resizebox{1.0\columnwidth}{!}{
\setlength{\tabcolsep}{0.5em}%
\begin{tabular}{c|ccccc}
\toprule
\rowcolor{lime!10}Learning Rate $\eta$ & $\frac{1}{255}$ & {*}$\frac{2}{255}$ & $\frac{3}{255}$ & $\frac{4}{255}$ & $\frac{5}{255}$\tabularnewline
\midrule
SRC $\taun$ & 0.404, 3.0 & 0.416, 3.1 & 0.394, 2.8 & 0.398, 2.8 & 0.387, 2.8\tabularnewline
\bottomrule
\end{tabular}
	}
\caption{Learning Rate $\eta$ Parameter Search for NES. Fashion-MNIST dataset, $N=\infty$, $k=5$, $\varepsilon=\frac{4}{255}$.}
	\label{tab:paranes2}
\end{table}

\begin{table}[h]
	\centering
	\resizebox{1.0\columnwidth}{!}{
		\setlength{\tabcolsep}{0.5em}%
		\begin{tabular}{c|ccccc}
			\toprule
\rowcolor{lime!10} $\sigma$ & $\varepsilon/$0.125 & $\varepsilon/$0.25 & {*}$\varepsilon/$0.5 & $\varepsilon/$1.0 & $\varepsilon/$2.0\tabularnewline
SRC $\taun$ & 0.403, 2.9 & 0.407, 2.9 & 0.416, 3.1 & 0.400, 2.9 & 0.382, 2.8\tabularnewline
\bottomrule
\end{tabular}
}
\caption{$\sigma$ Parameter Search for NES. Fashion-MNIST dataset, $N=\infty$, $k=5$, $\varepsilon=\frac{4}{255}$.}
\label{tab:paranes}
\end{table}

As shown in Tab.~\ref{tab:paranes2} and Tab.~\ref{tab:paranes}, the $\taun$ performance of NES peaks
at $\eta=2/255$ and $\sigma=\varepsilon/0.5$.
These results show that a small update step size and a relatively large
variance for the randomly generated samples are beneficial.
Specifically, we speculate that
a relatively small update step size allows the algorithm to adjust
the \emph{relative order} in a fine-grained manner, while
a relatively large variance is helpful
for the NES algorithm to escape from a local optimum.
Hence, we use this setting for
all the rest experiments with NES.

	\subsection{Simultaneous Perturbation Stochastic Approximation (SPSA)}

SPSA~\citep{spsa-atk} is based on \citet{spsa-algo}. In particular,
the implementation of \citep{spsa-atk} is very similar to the NES implementation
discussed above.
The only difference between the implementations of NES and SPSA lies in the
sampling of random vectors used for gradient estimation.
In \citep{spsa-atk}, the random vectors
are sampled from Rademacher distributions
(\ie, Bernoulli $\pm 1$) instead of the Gaussian distributions:
$\vz = \delta\vu = \delta[u_1,u_2,\ldots,u_D]$, and
$\forall i\in{1,\ldots,D},~ u_i \sim \text{Rademacher}()$, where $\delta$
is a tunable parameter controlling the infinite norm of random vectors.
Apart from that, we also set the batch size as $H=50$ for all SPSA experiments.

Compared to the NES algorithm in the experiments,
we speculate that such random vector sampling strategy endows SPSA
with better ability to jump out from local optimum due to a larger norm of
the random perturbations, then facilitates estimation of more effective gradients.
As a result, SPSA performs better than NES in some difficult cases.

\subsubsection{Parameter Search for SPSA}


\begin{table}[h]
\centering
\resizebox{1.0\columnwidth}{!}{
\setlength{\tabcolsep}{0.5em}%
\begin{tabular}{c|ccccc}
\toprule
\rowcolor{lime!10}Learning Rate $\eta$ & $\frac{1}{255}$ & {*}$\frac{2}{255}$ & $\frac{3}{255}$ & $\frac{4}{255}$ & $\frac{5}{255}$\tabularnewline
\midrule
SRC $\taun$ & 0.383, 3.0 & 0.407, 3.2 & 0.374, 2.8 & 0.360, 2.8 & 0.365, 2.8\tabularnewline
\bottomrule
\end{tabular}
	}
\caption{Learning Rate $\eta$ Parameter Search for SPSA. Fashion-MNIST dataset, $N=\infty$, $k=5$, $\varepsilon=\frac{4}{255}$.}
	\label{tab:paraspsa2}
\end{table}

\begin{table}[h]
	\centering
	\resizebox{1.0\columnwidth}{!}{
		\setlength{\tabcolsep}{0.5em}%
		\begin{tabular}{c|ccccc}
\toprule
\rowcolor{lime!10} Perturbation size $\delta$ & $\frac{1}{255}$ & {*}$\frac{2}{255}$ & $\frac{3}{255}$ & $\frac{4}{255}$ & $\frac{5}{255}$\tabularnewline
SRC $\taun$ & 0.322, 2.7 & 0.407, 3.2 & 0.401, 2.9 & 0.400, 2.9 & 0.397, 2.8\tabularnewline
\bottomrule
\end{tabular}
}
\caption{Perturbation Size $\delta$ Parameter Search for SPSA. Fashion-MNIST dataset, $N=\infty$, $k=5$, $\varepsilon=\frac{4}{255}$.}
\label{tab:paraspsa}
\end{table}


According to the parameter search in Tab.~\ref{tab:paraspsa2} and Tab.~\ref{tab:paraspsa}, we set
$\eta=2/255$ and $\delta=2/255$ as the default parameter in all other experiments
with SPSA.

\end{document}